\documentclass[sigconf]{acmart}

\AtBeginDocument{%
  }

\copyrightyear{2025}
\acmYear{2025}
\setcopyright{acmlicensed}\acmConference[KDD '25]{Proceedings of the 31st ACM SIGKDD Conference on Knowledge Discovery and Data Mining V.2}{August 3--7, 2025}{Toronto, ON, Canada}
\acmBooktitle{Proceedings of the 31st ACM SIGKDD Conference on Knowledge Discovery and Data Mining V.2 (KDD '25), August 3--7, 2025, Toronto, ON, Canada}
\acmDOI{10.1145/3711896.3736819}
\acmISBN{979-8-4007-1454-2/2025/08}

\theoremstyle{plain}

\theoremstyle{definition}

\theoremstyle{remark}

\usepackage{enumitem}

\usepackage{wrapfig}
\usepackage{graphicx}
\usepackage{subcaption}

\usepackage{diagbox}

\usepackage{multirow}
\usepackage{microtype}
\usepackage{tabu}

\newcolumntype{M}[1]{>{\centering\arraybackslash}m{#1}}

\usepackage{pifont}
\usepackage{adjustbox}
\usepackage{xspace}
\usepackage{xcolor}

\usepackage{enumitem}

\usepackage{pifont}

\definecolor{BrickRed}{rgb}{0.8, 0.25, 0.33}
\definecolor{ForestGreen}{rgb}{0.13, 0.55, 0.13}

\usepackage{tikz}

\definecolor{NavyBlue}{RGB}{0,127,255} 
\definecolor{myyellow}{RGB}{255,217,102} 

\newcommand{\LATER}[1]{}

\newcommand{\methodlong}{\textbf{\underline{d}}iffusion-based \textbf{\underline{i}}nterventional, \textbf{\underline{m}}\emph{}ulti-outcome \textbf{\underline{e}}stimation\xspace}
\newcommand{\Methodlong}{\textbf{\underline{D}}iffusion-Based \textbf{\underline{I}}nterventional, \textbf{\underline{M}}\emph{}ulti-Outcome \textbf{\underline{E}}stimation\xspace}
\newcommand{\methodshort}{DIME\xspace}

\begin{document}

\title{A Diffusion-Based Method for Learning the Multi-Outcome Distribution of Medical Treatments}

\author{Yuchen Ma}
\affiliation{%
  \institution{Munich Center for Machine Learning \& LMU Munich}
  \city{Munich}
  \country{Germany}}
\email{yuchen.ma@lmu.de}

\author{Jonas Schweisthal}
\affiliation{%
  \institution{Munich Center for Machine Learning \& LMU Munich}
  \city{Munich}
  \country{Germany}}
\email{jonas.schweisthal@lmu.de}

\author{Hengrui Zhang}
\affiliation{%
  \institution{University of Illinois, Chicago}
  \city{Chicago}
  \country{USA}}
  \email{hzhan55@uic.edu}

\author{Stefan Feuerriegel}
\affiliation{%
  \institution{Munich Center for Machine Learning \& LMU Munich}
  \city{Munich}
  \country{Germany}}
\email{feuerriegel@lmu.de}

\begin{abstract}
In medicine, treatments often influence multiple, interdependent outcomes, such as primary endpoints, complications, adverse events, or other secondary endpoints. Hence, to make optimal treatment decisions, clinicians are interested in learning the distribution of multi-dimensional treatment outcomes. However, the vast majority of machine learning methods for predicting treatment effects focus on single-outcome settings, despite the fact that medical data often include multiple, interdependent outcomes. To address this limitation, we propose a novel diffusion-based method called \methodshort to learn the joint distribution of multiple outcomes of medical treatments. Our \methodshort method addresses three challenges relevant in medical practice: (i)~our method is tailored to learn the joint interventional distribution of multiple medical outcomes, which enables reliable decision-making with uncertainty quantification rather than relying solely on point estimates; (ii)~our method explicitly captures the dependence structure between outcomes; and (iii)~our method can handle outcomes of mixed type, including binary, categorical, and continuous variables. In our method, we take into account the fundamental problem of causal inference, where only outcomes for the assigned treatment are observed, through causal masking. For training, our method decomposes the joint distribution into a series of conditional distributions with a customized conditional masking to account for the dependence structure across outcomes. For inference, our method auto-regressively generates predictions. This allows our method to move beyond point estimates of causal quantities and thus learn the joint interventional distribution. To the best of our knowledge, \methodshort is the first neural method tailored to learn the joint, multi-outcome distribution of medical treatments. Across various experiments, we demonstrate that our method effectively learns the joint distribution and captures shared information among multiple outcomes. 
\end{abstract}




\maketitle

\section{Introduction}
\label{sec:intro}

\begin{figure}[!t]
\begin{center}
\centerline{\includegraphics[width=1.1\columnwidth]{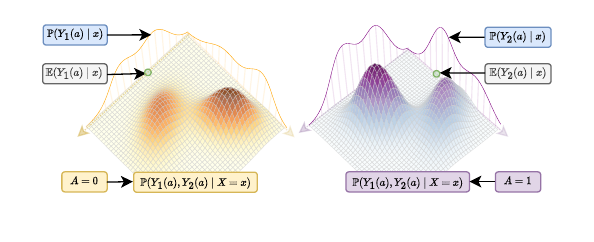}}
\vspace{-0.3cm}
\caption{Motivating example showing insights generated from the {joint} interventional distribution for an individual. Here, $Y^1$ and $Y^2$ are the two outcomes of interest for a patient with covariates $X = x$. Left: no treatment ($A=0$); Right: with treatment ($A=1$). We see three different causal quantities with different levels of insight. (1)~Point estimates are shown as green dots, which represent $\mathbb{E}[Y^i(a) \mid X =x]$ for $i \in\{1,2\}$. Yet, point estimates provide only very limited information, especially they lack any uncertainty quantification that is needed for reliable decision-making. (2)~The \emph{marginal} distribution of potential outcomes is shown as a line plot in the background. The marginal distribution is given by  $p(Y^i(a) \mid X = x)$. Yet, while the marginal distribution offers uncertainty quantification, it does \emph{not} capture the dependent structure between outcomes. (3)~The \emph{joint} interventional distribution of outcomes is shown as a 3D plot. It gives $p(Y^1(a), Y^2(a) \mid X = x)$ and, in contrast, thus provides detailed information about the dependence structure across outcomes in personalized medicine. Our focus is (3).}
\label{fig:motivating_example}
\end{center}
\vspace{-0.3cm}
\end{figure}

In personalized medicine, the goal is to predict the outcomes of medical treatments, so that clinicians can make optimal, patient-specific medical decisions \cite{Feuerriegel2024, Weberpals2025, doutreligne2025step}. Importantly, treatments typically influence multiple, interdependent outcomes. Hence, it is common practice in medicine to not only predict the primary outcome of a treatment (often called {efficacy endpoint}) but also secondary outcomes capturing complications or adverse events (often called {safety endpoints}).  Therefore, to support clinical decision-making, methods are needed that learn the full distribution of multi-dimensional treatment outcomes rather than focusing on single outcome predictions (see Fig.~\ref{fig:motivating_example}).

\vspace{0.1cm}
\textbf{Examples.} \emph{In oncology, an anti-cancer therapy may not only shrink the tumor size (primary outcome) but also cause side effects such as fatigue or immune suppression (adverse events) \cite{thai2021lung}. Similarly, in cardiology, a drug prescribed for heart failure might improve cardiac function as the primary outcome while also increasing the risk of complications such as hypotension~\citep{page2016drugs}. These outcomes, however, are often interdependent. For example, in oncology, a higher chemotherapy dosage may lead to greater tumor reduction but the higher toxicity also makes adverse reactions more likely~\citep{frei1980dose,feliu2020management}}
\vspace{0.1cm}

Machine learning has emerged as a powerful tool for predicting the outcomes of medical treatments \cite{Feuerriegel2024}. In many medical applications, a key estimand is the \emph{conditional average potential outcome} (CAPO), which is the expected outcome under a given treatment for a specific subgroup of patients, defined by their characteristics \cite{kunzel2019metalearners, ma2024diffpo}. However, treatments are \emph{interventions} and thus require a \emph{causal} approach to learn true cause-effect relationships \cite{Kern2025}. Yet, predicting CAPOs is inherently challenging due to the \emph{fundamental problem of causal inference} \cite{holland1986statistics, peters2017elements}: for any patient, we only observe the outcome of the treatment actually given, but we do not observe the counterfactual outcomes of treatments that were not administered. As a result, standard machine learning methods are biased if applied to causal questions \cite{shalit2017estimating, curth2021inductive} and can thus lead to wrong or harmful decisions. As a remedy, tailored methods are needed to correctly predict the outcomes of treatments.

So far, there is rich literature aimed at predicting the effect of treatments in \emph{single} outcome settings (e.g., \cite{chipman2010bart, johansson2016learning, shalit2017estimating, wager2018estimation, johansson2018learning, kunzel2019metalearners, nie2021quasi, curth2021inductive, curth2021nonparametric,kennedy2023towards}), yet despite the fact that medical data typically captures \emph{multiple} outcomes \cite{maag2021modeling, naumzik2023data, naumzik2024data}. There are only a few works (e.g., \cite{lin2000scaled, roy2003scaled, teixeira2011statistical, yoon2011alternative, kennedy2019estimating, yao2022efficient, wu2023blessings}) that focus on predicting treatment effects for multiple outcomes. However, these methods have clear \emph{limitations}. For example, they either rely on strong parametric assumptions \cite{lin2000scaled, roy2003scaled, teixeira2011statistical}, are limited to randomized controlled trials \cite{lin2000scaled, roy2003scaled, teixeira2011statistical}, require additional knowledge of the causal graph such as the causal structure between outcomes \cite{yoon2011alternative, yao2022efficient, wu2023blessings}, or focus on the average effect instead of individualized outcomes \cite{kennedy2019estimating}.

Another limitation of most machine learning methods for predicting treatment effects is that these typically focus on point estimates. However, point estimates fail in uncertain quantification and thus do \emph{not} allow for reliable decision-making in medicine \cite{heckman1997making, zampieri2021using, banerji2023clinical, kneib2023rage, Kern2025}. For example, a point estimate would say that, on average, a tumor shrinks by a certain amount, but it does not say how likely this is. In other words, the probability of a treatment success would be unclear to clinicians and patients, because it would require distributional information about the treatment effect. In contrast, only a few methods learn the distribution of outcomes \cite{melnychuk2023normalizing, ma2024diffpo}, yet typically only in \emph{single} outcome settings. In contrast, methods for learning the \emph{multi}-outcome distribution of medical treatments are lacking.

To fill this gap, we aim to \emph{learn the multi-outcome distribution of medical treatments}. For this, we propose a novel, diffusion-based method, which we call \methodshort (which is short for \methodlong). Our \methodshort method has three key advantages that are highly relevant to medical practice: \textbf{(i)}~\methodshort is tailored to learn the \emph{joint}, \emph{interventional} distribution of multiple CAPOs, which enables \emph{reliable} decision-making with uncertainty quantification rather than relying solely on point estimates; \textbf{(ii)}~\methodshort explicitly captures the \emph{dependence structure} between CAPOs; and \textbf{(iii)}~\methodshort can handle \emph{mixed-type outcomes}, including binary, categorical, and continuous variables. 

Methodologically, our \methodshort method takes into account the fundamental problem of causal inference using a causal masking step. For training, we decompose the joint distribution into a series of conditional distributions with customized conditional masking to account for dependence structure across outcomes. For inference, our method auto-regressively generates predictions. This allows our method to move beyond point estimates of causal quantities and thus learn the joint interventional distribution. To the best of our knowledge, \methodshort is the first neural method tailored to learn the joint, multi-outcome distribution of medical treatments.

Overall, our \textbf{main contributions} are the following:\footnote{Code is available at \url{https://github.com/yccm/DIME}.)} 
\begin{enumerate}[leftmargin=15pt]
\item We propose a tailored diffusion-based method that learns the joint interventional distribution of multiple outcomes. Thereby, our method paves the way for reliable, uncertainty-aware decision-making in personalized medicine.  
\item We design a novel way for effectively capturing latent interdependencies among multiple outcomes by decomposing the complex distribution into a series of simpler conditional distributions with customized conditional masking. 
\item We conduct extensive experiments across various medical settings, demonstrating that our method achieves state-of-the-art performance. 
\end{enumerate}

\section{Related Work}
\label{sec:related_work}

In the following, we review key literature streams relevant to our paper, namely, (i)~CAPO prediction with a single outcome; (ii)~treatment effects with multiple outcomes; and (iii)~learning the distribution of causal quantities. Note that there are clear gaps in the literature: methods in (i) and (ii) fail to learn distribution properties unlike our method, and the existing methods in (iii) are limited to single-outcome settings but extending them to multi-outcome settings is non-trivial and presents our contribution.

\subsection{CAPO prediction with a single outcome}

A growing number of machine learning methods have been developed to predict causal quantities -- most notably, the {conditional average treatment effect} (CATE) and CAPOs -- from observational data (e.g., \cite{chipman2010bart, johansson2016learning, shalit2017estimating, wager2018estimation, johansson2018learning, yoon2018ganite,kunzel2019metalearners, nie2021quasi, curth2021inductive, curth2021nonparametric,kennedy2023towards}). Although the task of predicting the CATE and CAPOs is asymptotically equivalent, many methods aim \emph{directly} at predicting CATE and, to improve estimatability, adopting a simplifying inductive bias whereby the CATE is assumed to be simpler than the CAPOs \cite{curth2021inductive}. Hence, while methods for predicting CATE are mathematically equivalent to predicting CAPOs (in infinite sample regimens), the former may lead to suboptimal performance when predicting CAPOs in finite sample regimens \cite{curth2021inductive}. Further, the aforementioned methods are designed to learn the effect of a treatment on a single outcome but are \textbf{not} designed for handling multiple outcomes.

\subsection{CAPO prediction with multiple outcomes}

Several works consider to estimate causal quantities -- such as treatment effects -- in multiple-outcome settings \cite{lin2000scaled, roy2003scaled, teixeira2011statistical, yoon2011alternative, yao2022efficient, wu2023blessings}. Here, the idea is to learn the shared information between outcomes and thus account for the dependence structure among multiple treatments. However, existing methods have clear \emph{limitations}. Some rely on strong parametric assumptions \cite{lin2000scaled, roy2003scaled, teixeira2011statistical}. Other methods are limited to randomized controlled trials \cite{lin2000scaled, roy2003scaled, teixeira2011statistical}, or require additional knowledge of the causal graph such as the causal structure between outcomes \cite{yoon2011alternative, yao2022efficient, wu2023blessings}. Because of these reasons, existing methods are \textbf{not} applicable to our setting. On top of that, these methods aim at a \emph{different} task: they learn point estimates and therefore neglect distribution information, which would be needed for uncertainty quantification and thus for reliable decision-making.

Closest to our work is the method in \cite{kennedy2019estimating}, which, unlike the above methods, builds upon a setting that is similar to ours but focuses on a \emph{different} objective. Specifically, the work \cite{kennedy2019estimating} introduces a nonparametric doubly robust method for estimating scaled treatment effects on multiple outcomes. However, this method is limited to causal inference for \emph{averaged} quantities, which are expressed through the (conditional) mean of potential outcomes. In contrast, our objective is to predict the \emph{individualized} outcomes, which offers a granular understanding of the treatment effect heterogeneity and thus allows one to personalize treatment decisions to individual patient profiles. Hence, relying on averaged causal quantities for medical decision-making can be misleading and, in some cases, even dangerous \cite{spiegelhalter2017risk, van2019communicating}.

\subsection{Learning the distribution of causal quantities}

Several methods have been developed to learn the \textit{distribution} of causal quantities. Examples include methods aimed at interventional density estimation \cite{melnychuk2023normalizing}, distributions of potential outcomes \cite{ma2024diffpo}, and distributional treatment effects \cite{byambadalai2024estimating}. However, these methods are restricted to single-outcome settings. The methods have additional limitations because of which they are not directly applicable to our setting. Some methods \cite{byambadalai2024estimating} are constrained to randomized experiments and are thus not designed to handle observational data. Other works \cite{melnychuk2023normalizing, byambadalai2024estimating} focus only on averaged rather than individualized outcomes. Hence, these methods do \textbf{not} provide suitable baselines for our objective.  

A closely related work is DiffPO \cite{ma2024diffpo}. However, the DiffPO method is again limited to single outcome settings. Hence, it fails to learn the shared information between different outcomes and thus fails to account for the dependence structure. Further, the DiffPO can not handle mixed-type outcomes (e.g., binary, categorical, continuous variables). Nevertheless, we later adapt \cite{ma2024diffpo} as one of our main baselines but find that is suboptimal due to the reasons discussed below. 

One may think that a na{\"i}ve approach is to extend a method from a single outcome to a multi-outcome setting by simply modeling each outcome independently. However, this approach fails to learn the \emph{joint} distribution, as it only captures the marginal distribution $p(Y^i(a) \mid X = x)$ for each outcome $Y^i$, without considering dependencies across outcomes. This is shown in Fig.~\ref{fig:motivating_example}. In contrast, our aim is to learn the full \emph{joint} interventional distribution $p(Y^1(a), Y^2(a) \mid X = x)$, which provides richer insights into how outcomes co-vary under the same treatment and thus allows to capture the dependence structure among different health outcomes in personalized medicine. We later use such a na{\"i}ve approach as our key baseline and to show why our task is non-trivial.

\textbf{Research gap:} As discussed above, there is \textbf{no} method for learning the distributions of multiple CAPOs. To fill this gap, we propose our method for \methodlong called \methodshort. To the best of our knowledge, ours is the first neural method specifically tailored to learn the joint distribution of multiple, interdependent outcomes of a medical treatment.

\section{Preliminaries on Diffusion Models}
\label{sec:pre-diff}

Score-based diffusion models (also known as denoising diffusion probabilistic models) \cite{sohl2015deep, song2019generative, ho2020denoising, song2020score} are likelihood-based generative models that have demonstrated remarkable success in modeling complex data distributions. These models operate by gradually perturbing data through a forward diffusion process and then learning to reverse this process in order to generate samples from the learned  data distribution.

Score-based diffusion models begin with an unknown data distribution $p_0$. For an initial sample $x_0 \sim p_0$, the forward diffusion process is then applied to this sample by gradually adding noise to $x_0$. Let $ x_t $ denotes the state at time $ t \in [0, T] $. As time increases from $0$ to $T$, the sample $x_t$ becomes increasingly noisy until its distribution $p_t$ eventually is driven toward a simple and tractable terminal distribution $ p_T $ (typically Gaussian). Formally, the forward diffusion process is described by the stochastic differential equation (SDE)
\begin{equation}
\label{eq:SDE}
    \mathrm{d}x = f(x,t)\,\mathrm{d}t + g(t)\,\mathrm{d}w,
\end{equation}
where $ f:\mathbb{R}^d \times [0,T] \to \mathbb{R}^d $ is the drift term, $ g:[0,T] \to \mathbb{R}_{+} $ is the diffusion coefficient, and $ w(t) $ is a standard Wiener process. The time-evolving density $ p_t(x) $ satisfies the corresponding Fokker–Planck equation.

To sample from $ p_0 $, one simulates the reverse-time SDE, which, under suitable regularity conditions, is given by
\begin{equation}
    \mathrm{d}x = \left[f(x,t) - g(t)^2\,\nabla_x \log p_t(x)\right] \mathrm{d}t + g(t)\,\mathrm{d}\overline{w},
\end{equation}
where $ \overline{w}(t) $ is a reverse-time Wiener process. The term $ \nabla_x \log p_t(x) $ is the score function, i.e., the gradient of the log-density of the perturbed data at time $ t $. The reverse SDE is similar to the forward one but contains an extra term involving the score function.

In practice, the $ \nabla_x \log p_t(x) $ is unknown and not available in closed form. Thus, the score function is approximated by a time-dependent neural network $ s_{\theta}(x,t) $, with parameters $ \theta $, trained using score matching objectives. Once an accurate approximation of $ \nabla_x \log p_t(x) $ is obtained, it can be used to simulate the reverse SDE, effectively transforming a simple noise sample drawn from $p_T$ back into a sample from the target distribution $p_0$.

\section{Problem Formulation}\label{sec:problem_setup}

\textbf{Notation:} We use capital letters to denote random variables and small letters for their realizations. We denote the propensity score as $\pi(x)=\mathrm{Pr}(A=1 \mid X=x)$, which gives the treatment assignment mechanism in observational data. 

\textbf{Setting:} We consider a setting with \emph{multiple} outcomes of interest for a medical treatment (e.g., blood pressure and heart rate for a cardiovascular drug). At training time, we have information about observed confounders $X \in$ $\mathcal{X} \subseteq \mathbb{R}^{d_x}$, a treatment $A \in \mathcal{A} = \{0,1\}$, and $k$ outcomes $Y^1, Y^2, \dots, Y^k$. We consider multi-type outcomes. That is, for each outcome $i \in \{1, \dots, k\}$, $Y^i \in \mathcal{Y}^i$, we have $\mathcal{Y}^i \subseteq \mathbb{R}$ if $Y^i$ is continuous, and $\mathcal{Y}^i = \{1,2,\dots,L^i\}$ if $Y^i$ is categorical. We write $\tilde{Y} = \bigl(Y^1, Y^2, \dots, Y^k\bigr)^\top \in \mathcal{Y}$ to denote the joint vector of all outcomes, where the outcome space is given by $\mathcal{Y} = \mathcal{Y}^1 \times \mathcal{Y}^2 \times \cdots \times \mathcal{Y}^k$. We have access to an observational dataset $\mathcal{D}=\left(x_j, a_j, \tilde{y}_j\right)_{j=1}^n$ sampled i.i.d. from the unknown joint distribution $p(X,A,\tilde{Y})$.

Our setting is highly relevant to clinical practice. For instance, in critical care, the patient covariates $X$ may consist of different risk factors (e.g., age, gender, prior diseases), the treatment $A$ can be whether a certain drug is taken, and the outcomes $\tilde{Y}$ describe a patient's health along different multiple clinical measurements (e.g., blood pressure, heart rate, blood oxygen saturation, and respiratory frequency). To allow for uncertainty quantification rather than point estimates, we are interested in learning the interventional distribution of outcomes.

We build upon the Neyman-Rubin potential outcomes framework \cite{rubin2005causal}. The joint vector of \emph{potential outcomes} (PO) is defined as $\tilde{Y}(a) = \bigl(Y^1(a), Y^2(a), \dots, Y^k(a)\bigr)^\top \in \mathcal{Y}$, where $Y^i(a)$ denotes the $i$-th PO under the treatment intervention $A = a$. Due to the fundamental problem of causal inference \cite{holland1986statistics}, only one of the POs is observed for each individual, i.e., either $\tilde{Y}(1)$ if the treatment is administered ($A=1$), or $\tilde{Y}(0)$ if not treated ($A=0$). Consequently, the observed outcome is given by $\tilde{Y}=A \tilde{Y}(1)+\left(1-A\right) \tilde{Y}(0)$.

For each outcome $Y^i$, the \emph{expected conditional average potential outcome (CAPO)} is given by $\mathbb{E}[Y^i(a) \mid X=x]$, which is the expected outcome under a treatment assignment (intervention) $A = a$ for an individual with covariates $X=x$. Likewise, the CATE for outcome $Y^i$ is given by $\tau^i(x)=\mathbb{E}[Y^i(1)-Y^i(0) \mid X=x]$, which is the expected treatment effect for an individual with covariate values $X=x$. Identifiability is ensured by the standard assumptions in the PO framework (i.e., consistency, unconfoundedness, and overlap), which is analogous to previous literature (e.g., \cite{curth2021inductive,curth2021nonparametric,kennedy2023towards}).

\textbf{Target estimand:} The target estimand is the joint interventional distribution 
\begin{equation}
    p\left(Y^1(a), Y^2(a), \ldots, Y^k(a) \;\middle|\; X=x\right) ,
\end{equation}
which is equivalent to $p\bigl(\tilde{Y}(a) \in B \mid X = x\bigr)$ for any measurable set $B \subseteq \mathcal{Y}$. When outcomes are mixed-type, the corresponding density/mass function, denoted by $p_{\tilde{Y}\mid X,A}(y^i \mid y^v, x, a)$, must be interpreted as a density (with respect to the Lebesgue measure) for continuous outcomes and a probability mass function for categorical outcomes. Our interest is in learning the full joint interventional distribution $p\left(Y^1(a), Y^2(a), \ldots, Y^k(a) \;\middle|\; X=x\right)$ with treatment $A=a$, which captures the complex dependent relationships among the multiple outcomes. 

In medical decision-making, the \emph{distribution} of potential outcomes is often more informative than a single point estimate. The reason is that clinicians want to understand the uncertainty associated with how patient outcomes develop in response to treatments rather than rely only on the most likely value \cite{heckman1997making,kneib2023rage}. Consequently, learning the full outcome distribution provides clinical insights into the possible range of potential clinical trajectories, which, in turn, helps in personalized decision-making.

\textbf{\emph{Why is our task non-trivial?}} A na{\"i}ve approach for handling multiple outcomes is to model each outcome independently. Yet, this would capture only the marginal distribution $p(Y^i(a) \mid X = x)$. Although straightforward, such na{\"i}ve approach ignores potential dependencies and shared information among outcomes -- which is common in medicine, where outcomes are often co-occur and are thus correlated or somehow interdependent (for example, weight loss medications such as GLP-1 simultaneously affect both body mass index and blood pressure)~\citep{katout2014effect}. However, this would not \textbf{not} yield the full \emph{joint} interventional distribution (see Fig.~\ref{fig:motivating_example}). The latter is given by $p(Y^1(a), Y^2(a) \mid X = x)$. While one can obtain the marginal distribution from the joint interventional distribution, the opposite is difficult. To see this, the marginal distributions of the potential outcome $Y^i$ can be obtained from the joint distribution via integration, i.e.,
\begin{equation*}
\hspace{-0.4cm}
    p\Bigl(Y^i(a) \;\Bigm|\; X=x \Bigr)=
    \int_{\Pi_{j \neq i} Y^j} p \Bigl( Y^1(a), \ldots, Y^k(a) \;\Bigm|\; X=x \Bigr) \;\mathrm{d} Y^{-i},
\hspace{-0.2cm}
\end{equation*}
where the differential $\mathrm{d} Y^{-i}$ represents the product measure over all variables $Y^j(a)$ for $j \neq i$. However, the reverse transformation (i.e., obtaining the joint from the marginals) is not possible without imposing strong independence assumptions. Hence, we later use the above na{\"i}ve as a baseline, and to address its shortcomings, we develop a tailored neural method in the following.

\section{\Methodlong (\methodshort)}

\subsection{Overview}

\begin{figure*}[!t]
\begin{center}
\centerline{\includegraphics[width=2.1\columnwidth]{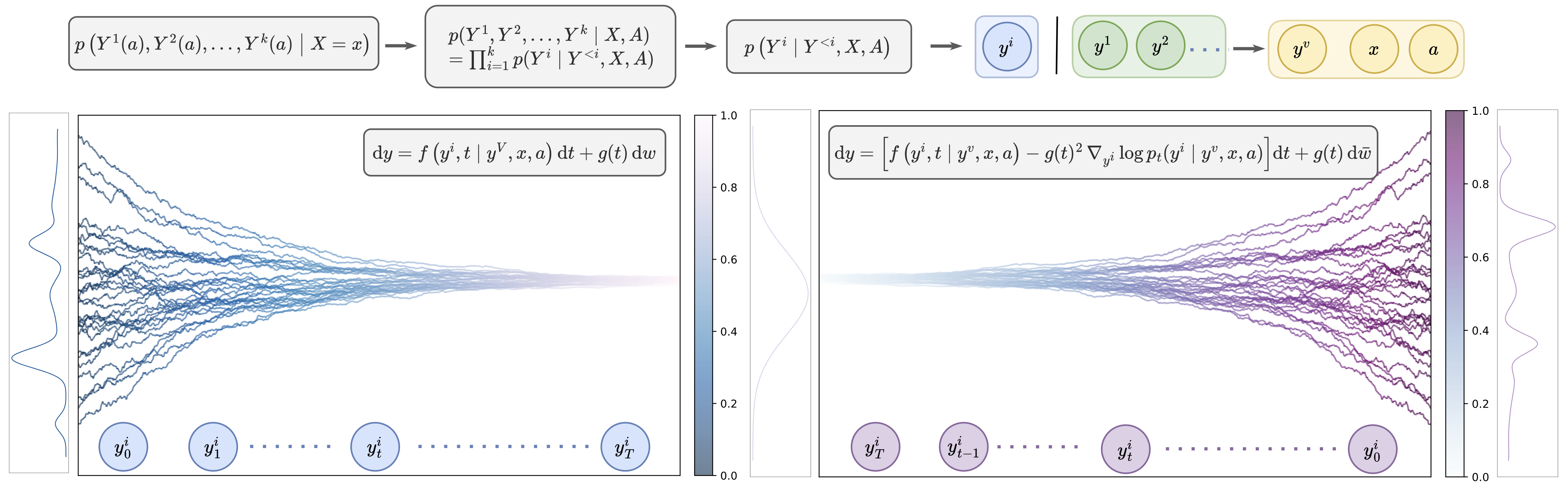}}
\caption{Overview of our \methodshort. We decompose the joint interventional distribution into a series of conditional distributions during training, which are learned by tailored score-based diffusion models with conditional masking.}
\label{fig:overview}
\vspace{-0.3cm}
\end{center}
\end{figure*}

In this section, we introduce our novel method for learning the joint interventional distribution of multiple outcomes called \methodshort. The overview of our method is in Fig.~\ref{fig:overview}. For training, we decompose the joint distribution into a series of simpler, more tractable conditional distributions (Section~\ref{sec:auto-gen}). These conditional distributions are then learned using tailored conditional score-based diffusion models, augmented with a {causal masking} to address the fundamental problem of causal inference and a {counterfactual masking} to ensure the dependencies between outcomes are accurately captured (Section~\ref{sec:csbdm}). For inference, our method generates samples from the learned conditional distributions in an autoregressive way, which enables accurate simulation of the interventional joint distribution.

Our \methodshort method has three strengths: \textbf{(i)}~\methodshort learns the \emph{interventional} distribution, which is challenging due to the fundamental problem of causal inference. We address this through a dedicated \emph{causal masking} step, which ensures that our method correctly handles the fact that only factual outcomes are observed for each treatment assignment, while counterfactual outcomes are unobserved. \textbf{(ii)}~\methodshort explicitly models \textit{joint} interventional distribution. Directly learning such a high-dimensional distribution is both computationally and statistically challenging, particularly because outcomes often exhibit interdependencies. To address this, we employ a decomposition strategy together with a \emph{counterfactual masking} mechanism that accounts for the dependence structure between different outcomes. \textbf{(iii)}~\methodshort can handle \emph{mixed-type} outcomes, including binary, categorical, and continuous variables. Such mixed-type data introduces additional complexity. Existing diffusion-based methods for single CAPOs \citep{ma2024diffpo} are typically limited to continuous variables. In contrast, \methodshort offers a unified method for modeling different outcome types.

\subsection{Decomposition for training and autoregressive generation for inference} 
\label{sec:auto-gen}

\subsubsection{Decomposition via chain rule}

Directly modeling the high-dimensional joint distribution $ p(Y^1, Y^2, \dots, Y^k \mid X=x, A=a) $ is generally computationally intractable. To address this, we decompose the joint distribution into a product of simpler conditional distributions using the chain rule of probability, i.e.,
\begin{equation}
p(Y^1, Y^2, \ldots, Y^k \mid X, A) = \prod_{i=1}^k p(Y^i \mid Y^{<i}, X, A),
\label{eq:factorization}
\end{equation}
where $ Y^{<i} = (Y^1, \dots, Y^{i-1}) $. This decomposition preserves dependencies among outcomes while making the learning process tractable. The factorization is not unique and depends on the ordering of outcomes. For example, an alternative factorization for a permutation $ \sigma $ of $ \{1, 2, \dots, k\} $ is
\vspace{-0.2cm}
\begin{multline}
p(Y^{\sigma(1)} \mid X, A) \cdot p(Y^{\sigma(2)} \mid Y^{\sigma(1)}, X, A) \\
\cdots p(Y^{\sigma(k)} \mid Y^{\sigma(1)}, \dots, Y^{\sigma(k-1)}, X, A).
\label{eq:alternative_factorization}
\end{multline}
We later train our method over $\Sigma$ different possible orderings $\sigma$, so that our method is robust to the choice of factorization. Generally, it is preferred to learn overall possible orderings but, for high-dimensional settings, it may be preferable to limit $\Sigma$ for computational reasons.

\subsubsection{Training strategy using causal masking and conditional masking}

To learn the conditional interventional distributions in a structured manner, we adopt a hierarchical training strategy. 

Our \methodshort first learns the marginal conditional distributions $p(Y^i \mid X, A) $ for each $ i \in \{1, 2, \dots, k\} $.  It then proceeds to learn higher-order conditional distributions $p(Y^i \mid Y^j, X, A) $ for $ i \neq j $, followed by $p(Y^i \mid Y^j, Y^k, X, A) $ for $ i \neq j \neq k $, and so on. By incrementally incorporating dependencies, such a hierarchical approach ensures that our method captures increasingly complex dependencies among outcomes. Formally, for any subset of indices $v \subseteq\{1, \ldots, i-1\}$, we have $Y^v=\left\{Y^i: i \in v\right\}$. Then, the training process learns conditional distribution $p(Y^i \mid Y^v, X, A)$ and thus progressively captures dependencies among outcomes.

To ensure that our \methodshort correctly accounts for the causal structure and thus for the fundamental problem of causal inference, we introduce binary masks defined with regard to the outcome index set $\{1, \ldots, k\}$, which we call \textbf{\emph{causal masking}}. The causal masking consists of two masks: 
\begin{enumerate}[leftmargin=15pt]
\item \emph{Input masking:} We define an input mask $m_{\mathrm{i}} \in \{0,1\}^k$ that indicates the positions corresponding to observed outcomes in the training data. Due to the fundamental problem of causal inference \cite{holland1986statistics} (as described in Sec.~\ref{sec:problem_setup}), only one potential outcome $\tilde{Y}(a)$ is observed per individual. During training, this input mask informs the method of which entries are available in the observational dataset. 
\item \emph{Target masking:} A target mask $m_{\mathrm{t}} \in \{0,1\}^k$ is applied to indicate the positions where the loss is computed. During training, only the outcome components undergoing denoising at a particular step contribute to the training loss. At inference, all the outcomes are predicted. 
\end{enumerate}

We further apply a \textbf{\emph{conditional masking}} to better learn the conditional distributions in Eq.~\ref{eq:factorization}. Recall that the prediction is not only conditioned on $x$ and $a$  but on a certain outcome $y^i$ as well. We thus design special conditional masking for our hierarchical training. For a given permutation $\sigma$, let $m_{\mathrm{c}}^\sigma \in \{0,1\}^k$ denote the conditional mask. The conditional masking specifies which outcome components are used as conditioning information in a given conditional distribution. This is crucial when different orderings yield different sets $Y^{<i}$ for conditioning, so that our method can be trained on multiple orderings $\sigma$.

\subsubsection{Autoregressive generation at inference}
\label{sec:auto-inference}

During inference, outcomes are sampled autoregressively based on the learned conditional distributions. Specifically, for a given ordering $\sigma$, our method first samples $Y^{\sigma(1)} \sim p\left(Y^{\sigma(1)} \;\middle|\; X=x, A=a\right)$, conditional on a covariate $x$ and treatment $a$. We then sequentially sample
\begin{equation}
Y^{\sigma(i)} \sim p \left(Y^{\sigma(i)} \;\middle|\; Y^{\sigma(1)}, \ldots, Y^{\sigma(i-1)}, X=x, A=a \right) ,
\end{equation}
for $i = 2, \ldots, k$, until all outcomes are generated. Because the factorization is not unique, we {aggregate} predictions across multiple orderings $\sigma$ by averaging over them. The rationale is that the aggregation should improve the robustness and ensure that dependencies among outcomes are captured, regardless of the specific factorization.

\subsection{Conditional score-based diffusion model for learning the distribution $p(Y^i \mid Y^{<i}, X, A)$}
\label{sec:csbdm}

In this section, we describe the conditional score-based diffusion model which we use for learning conditional distribution $p(Y^i \mid Y^{<i}, X, A)$ and which is the key component in the factorization of Eq.~\eqref{eq:factorization}. Recall that we adopt a hierarchical training strategy. Thus, in the intermediate steps, we learn the conditional distribution $p\left(Y^i \mid Y^v, X, A\right)$, where $v$ is the fixed index set from a subset of indices $\{1, \ldots, i-1\}$. To do so, we construct a conditional diffusion process that progressively perturbs samples from the target distribution into a tractable noise distribution, and then learn to reverse this process via score matching. Unlike a standard diffusion model, both our diffusion process and score matching explicitly account for the additional conditioning on previously generated outcomes $Y^{<i}$, besides conditioning on $X=x$ and $A=a$. 

In the following, we describe (i)~the forward diffusion process that gradually transforms a sample from the target distribution into a tractable noise distribution, (ii)~the reverse diffusion process that inverts this transformation by incorporating the (conditional) score function, (iii)~the training objective based on conditional score matching, and (iv)~the sampling procedure.

\subsubsection{Forward diffusion process.}

The forward diffusion process is over a time interval $t \in[0, T]$ that gradually perturbs a sample into a noise variable. We denote the initial sample as $y_0^i$, and the conditional density $p\left(y^i \mid y^v, x, a\right)$ as $p_0$, given an observational sample $(y^i, y^v, x, a)$. The forward diffusion process is then applied to this sample by gradually adding noise to $y_0^i$. We denote $y_t^i$ as the state of $y^i$ at time $t$.

We define a forward stochastic diffusion process by the stochastic differential equation
\begin{equation}
    \label{eq:forward_sde_condi}
    \mathrm{d}y = f\left(y^i, t \mid y^v, x, a\right)\, \mathrm{d}t + g(t)\, \mathrm{d}w,
\end{equation}
where $f\left(y^i, t \mid y^v, x, a\right)$ is a drift coefficient that depends on $y^i$, $t$, and the conditioning variables $y^v, x, a$; $g(t)$ is a diffusion coefficient that depends only on time $t$, and $w$ is a standard Wiener process.

The design of $f$ and $g$ is typically such that, at $t=0$, the distribution is $p_0\left(y^i \mid y^v, x, a\right)$ and, as $t \rightarrow T$, the process converges to a simple, tractable distribution $p_T\left(y^i \mid y^v, x, a\right)$ (e.g., a Gaussian distribution).

\subsubsection{Reverse diffusion process.}  

To generate samples from the target conditional, density $p_0\left(y^i \mid y^v, x, a\right)$, we need to reverse the forward diffusion. The reverse-time SDE is given by
\begin{equation}
    \label{eq:reverse_sde_condi}
    \mathrm{d}y = \Bigl[f\left(y^i, t \mid y^v, x, a\right) - g(t)^2\, \nabla_{y^i} \log p_t(y^i \mid y^v, x, a)\Bigr] \mathrm{d}t + g(t)\, \mathrm{d}\bar{w},
\end{equation}
where $\bar{w}$ is a reverse-time Wiener process and  $\nabla_{y^i} \log p_t(y^i \mid y^v, x, a)$ denotes the \emph{conditional score function}, i.e., the gradient (with respect to $y^i$) of the log-density at time $t$.

\subsubsection{Conditional score matching.}  

In practice, the exact conditional score $\nabla_{y^i} \log p_t(y^i \mid y^v, x, a)$ is generally intractable, because of which we approximate it using a neural network $s_{\theta}(y^i, t \mid y^v, x, a)$ parameterized by $\theta$. To train $s_{\theta}$, we adopt a denoising score matching objective that minimizes the weighted mean-squared error between the network output and the true score. Formally, the training loss is defined as
\begin{multline}
    \label{eq:score_matching_loss_condi}
    \mathcal{L}(\theta) = \mathbb{E}_{t,\,y_0^i,\,y_t^i\,y^v \,x,\,a} \Bigl[ \lambda(t)\, \Bigl\| s_{\theta}(y_t^i, t \mid y^v, x, a) - \\
    \nabla_{y_t^i} \log p_t(y_t^i \mid y^i_0, y^v, x, a) \Bigr\|^2 \Bigr],
\end{multline}
where $\lambda(t)$ is a weighting function (typically chosen proportional to $g(t)^2$) that balances the loss across time. In practice, the expectation is approximated by sampling time $t$, initial data $y_0^i$, and subsequently generating $y_t^i$ via the forward SDE in Eq.~\eqref{eq:forward_sde_condi}. The neural network $s_{\theta}$ is designed to condition only on the available variables by employing causal masking in the network architecture, thereby enforcing the appropriate conditional independence structure during training.

\subsubsection{Sampling procedure}  

After training, we generate samples from $p\left(y^i \mid y^v, x, a\right)$ by numerically integrating the reverse-time SDE in Eq.~\eqref{eq:reverse_sde_condi}. Specifically, we initialize the process at time $T$ with a sample $y_t^i \sim p_T$, where $p_T$ is a tractable prior distribution (e.g., a Gaussian). The reverse SDE in Eq.~\eqref{eq:reverse_sde_condi} is then integrated (using, e.g., an Euler-Maruyama scheme) from $t=T$ down to $t=0$, which yields a sample $y_0^i$ that approximates a draw from $p\left(y^i \mid y^v, x, a\right)$.

\subsection{Mixed-type loss} 

A key strength of our \methodshort is that we consider mixed-type outcomes (binary, categorical, and continuous). However, diffusion models are generally designed only for continuous data \cite{zhang2023mixed, lee2023codi, zhang2024diffusion}. As a remedy, when the outcome $y^i$ is categorical, there is no differentiable density over the finite set of possible categories. Thus, rather than using a continuous score-matching objective, we rely on a cross-entropy-based loss between the predicted distribution and the true one-hot label corresponding to the original category: 
\begin{equation}
\mathcal{L}(\theta)=\mathbb{E}_{y^i, y^v, x, a}\left[-\log p_\theta\left(y^i \mid  y^v, x, a\right)\right]
\end{equation}
where $p_\theta\left(y^i \mid  y^v, x, a\right)$ is the method's predicted probability distribution over the $L^i$ categories. 

Baseline methods struggle to model joint distributions for mixed-type outcomes (e.g., continuous, binary, and categorical variables). Mixed-type outcomes introduce additional complexity due to their fundamentally different properties: continuous and categorical variables lie in different spaces (e.g., $\mathbb{R}$ vs. a finite set).  Existing methods for CAPOs are limited to continuous variables (e.g., as is the case for DiffPO \cite{ma2024diffpo}) and thus cannot handle mixed-type outcomes. In contrast, our method explicitly handles mixed-type outcomes within a unified framework. We employ distinct continuous and discrete encoders to transform different types of variables into embeddings of identical dimensionality. Different feature types are mapped into a shared embedding space through our unique embedding design.

\subsection{Implementation Details}
\label{sec:imple-details}

We employ different continuous and discrete encoders to transform different feature types into embeddings of identical dimensionality. Both feature types are mapped into a shared embedding space: continuous features pass through a two-layer MLP with SiLU activations, and discrete features are assigned learnable embeddings. These embeddings (with positional encoding) are processed by six transformer blocks (each with four attention heads, following a ViT-inspired backbone) to capture complex interactions. Our denoising network employs a four-layer MLP-based architecture. It converts input data, conditional information, and sinusoidally encoded time steps into a common embedding space through linear projections and SiLU activations. For continuous outcomes, the network outputs the score directly; for categorical outcomes, a softmax layer predicts the probability distribution over categories. We use Adam (learning rate $=1 \times 10^{-3}$) for training. The implementation details are in Appendix~\ref{app:implentation_details_ours}.

\section{Experiments}
\label{sec:experiment}

\subsection{Setup}
\label{sec:exp-setup}

\subsubsection{Evaluation metrics for distributions} 

To evaluate the performance of our method in learning the distributions\footnote{While the full distribution of individual treatment effects is \emph{non-identifiable} \cite{melnychuk2024quantifying}, we emphasize the interventional distribution of the potential outcomes is identifiable. Hence, our evaluation focuses on the joint interventional distribution of multiple outcomes.}, we follow prior literature (e.g.,\cite{melnychuk2023normalizing, ma2024diffpo}), we use two standard metrics: the Wasserstein distance and the Kullback–Leibler (KL) divergence. Thereby, we directly measure the quality of the learned distribution and thus allow for a more comprehensive assessment than point-based error measures. Both are defined as follows:
\begin{enumerate}[leftmargin=15pt]
\item \emph{Wasserstein distance.} Given two probability measures $\nu_1$ and $\nu_2$ over $\mathcal{Y}$, the $p$-Wasserstein distance between $\nu_1$ and $\nu_2$ is defined as
\begin{equation}\label{eq:wasserstein}
    W_p(\nu_1, \nu_2) = \left( \inf_{\gamma \in \Gamma(\nu_1, \nu_2)} \int_{\mathcal{Y}\times \mathcal{Y}} \|\omega^1 - \omega^2\|^p \, \mathrm{d}\gamma(\omega^1, \omega^2) \right)^{\frac{1}{p}},
\end{equation}
where $\|\cdot\|$ denotes the Euclidean norm in $\mathbb{R}^k$, $\omega$ is the element drawn from the support of the distribution $\nu$, $p \geq 1$, $\Gamma(\nu_1, \nu_2)$ is the set of all couplings (i.e., joint measures) with marginals $\nu_1$ and $\nu_2$. For empirical evaluation of distributions, we have $n$ independent samples  $\{\xi_1, \xi_2, \ldots, \xi_n\} \sim \nu_1$ and $\{\eta_1, \eta_2, \ldots, \eta_n\} \sim \nu_2$; the empirical $p$-Wasserstein distance is then given by
\begin{equation}\label{eq:wasserstein_empirical}
    W_p(\nu_1, \nu_2) = \left( \min_{\sigma \in S_n} \frac{1}{n} \sum_{i=1}^{n} \|\xi_i - \eta_{\sigma(i)}\|^p \right)^{\frac{1}{p}},
\end{equation}
where $S_n$ denotes the set of all permutations of $\{1,2,\ldots,n\}$. In particular, for $p=1$ (in our experiments, we take $p=1$) the empirical Wasserstein distance reduces to
\begin{equation}\label{eq:wasserstein_empirical_p1}
    W_1(\nu_1, \nu_2) = \min_{\sigma \in S_n} \frac{1}{n} \sum_{i=1}^{n} \|\xi_i - \eta_{\sigma(i)}\|.
\end{equation}
\item \emph{KL divergence.} Given two probability measures $\nu_1$ and $\nu_2$ over $\mathcal{Y}$ with associated density functions $p$ and $q$, respectively. Then, the KL divergence between $\nu_2$ and $\nu_1$ is defined as
\begin{equation}
    D_{\mathrm{KL}}\left(\nu_1 \| \nu_2\right)=\int_{\mathcal{Y}} p(\omega) \log \frac{p(\omega)}{q(\omega)} \mathrm{d} y .
\end{equation}
Given $n$ samples $\left\{\xi_i\right\}_{i=1}^n \sim \nu_1$, the empirical divergence can be computed via
\begin{equation}
D_{\mathrm{KL}}\left(\nu_1 \| \nu_2\right) \approx \frac{1}{n} \sum_{i=1}^n \log \left(\frac{\hat{p}\left(\xi_i\right)}{\hat{q}\left(\xi_i\right)}\right)
\end{equation}
where $\hat{p}$ and $\hat{q}$ are estimates of the density functions $p$ and $q$, respectively.
\end{enumerate}

\begin{table*}[h]
\centering
\caption{Results showing in- \& out-of-sample empirical Wasserstein distance (i.e., $\hat{W}_1^{\text{in}}$ and $\hat{W}_1^{\text{out}}$) for different intervention ($a = 0$ and $a = 1$) on ACIC, IST and MIMIC datasets. Reported: mean $\pm$ standard deviation over ten-fold train-test splits.}
\label{tab:results_wass}
\resizebox{0.95\textwidth}{!}{%
\begin{tabular}{l|rr|rr|rr|rr|rr|rr}
\toprule
{} & \multicolumn{4}{c|}{ACIC} & \multicolumn{4}{c|}{IST} & \multicolumn{4}{c}{MIMIC} \\
\cmidrule{2-5}\cmidrule{6-9}\cmidrule{10-13}
{} & \multicolumn{2}{c|}{$a = 0$} & \multicolumn{2}{c|}{$a = 1$} & \multicolumn{2}{c|}{$a = 0$} & \multicolumn{2}{c|}{$a = 1$} & \multicolumn{2}{c|}{$a = 0$} & \multicolumn{2}{c}{$a = 1$} \\
{} &  \multicolumn{1}{c}{$\hat{W}_1^\text{in}$} &  \multicolumn{1}{c|}{$\hat{W}_1^\text{out}$} &   \multicolumn{1}{c}{$\hat{W}_1^\text{in}$} &  \multicolumn{1}{c|}{$\hat{W}_1^\text{out}$} &  \multicolumn{1}{c}{$\hat{W}_1^\text{in}$} &  \multicolumn{1}{c|}{$\hat{W}_1^\text{out}$} &   \multicolumn{1}{c}{$\hat{W}_1^\text{in}$} &  \multicolumn{1}{c|}{$\hat{W}_1^\text{out}$} & \multicolumn{1}{c}{$\hat{W}_1^\text{in}$} &  \multicolumn{1}{c|}{$\hat{W}_1^\text{out}$} &   \multicolumn{1}{c}{$\hat{W}_1^\text{in}$} &  \multicolumn{1}{c}{$\hat{W}_1^\text{out}$}\\
\midrule
        S-Net$^*$ \cite{kunzel2019metalearners} & 2.92 $\pm$ \tiny{0.22} & 3.18 $\pm$ \tiny{0.25} & 1.98 $\pm$ \tiny{0.17} & 2.22 $\pm$ \tiny{0.20} & 4.38 $\pm$ \tiny{0.32} & 4.62 $\pm$ \tiny{0.34} & 2.92 $\pm$ \tiny{0.23} & 3.18 $\pm$ \tiny{0.26} & 4.18 $\pm$ \tiny{0.30} & 4.42 $\pm$ \tiny{0.33} & 2.52 $\pm$ \tiny{0.20} & 2.75 $\pm$ \tiny{0.22} \\
        T-Net$^*$ \cite{kunzel2019metalearners} & 2.87 $\pm$ \tiny{0.21} & 3.12 $\pm$ \tiny{0.24} & 1.92 $\pm$ \tiny{0.18} & 2.15 $\pm$ \tiny{0.19} & 4.32 $\pm$ \tiny{0.31} & 4.56 $\pm$ \tiny{0.33} & 2.86 $\pm$ \tiny{0.22} & 3.12 $\pm$ \tiny{0.25} & 4.12 $\pm$ \tiny{0.29} & 4.35 $\pm$ \tiny{0.32} & 2.44 $\pm$ \tiny{0.19} & 2.68 $\pm$ \tiny{0.21} \\
        GANITE$^*$ \cite{yoon2018ganite} & 3.13 $\pm$ \tiny{0.54} & 3.42 $\pm$ \tiny{0.57} & 2.25 $\pm$ \tiny{0.43} & 2.48 $\pm$ \tiny{0.22} & 5.66 $\pm$ \tiny{0.34} & 6.88 $\pm$ \tiny{0.36} & 3.15 $\pm$ \tiny{0.49} & 3.42 $\pm$ \tiny{0.48} & 6.49 $\pm$ \tiny{0.42} & 6.68 $\pm$ \tiny{0.35} & 2.75 $\pm$ \tiny{0.52} & 2.98 $\pm$ \tiny{0.54} \\
        TARNet$^*$ \cite{shalit2017estimating} & 2.78 $\pm$ \tiny{0.20} & 3.05 $\pm$ \tiny{0.23} & 1.55 $\pm$ \tiny{0.16} & 1.64 $\pm$ \tiny{0.18} & 4.25 $\pm$ \tiny{0.30} & 4.48 $\pm$ \tiny{0.32} & 2.78 $\pm$ \tiny{0.21} & 3.05 $\pm$ \tiny{0.24} & 4.05 $\pm$ \tiny{0.28} & 4.19 $\pm$ \tiny{0.31} & 2.38 $\pm$ \tiny{0.18} & 2.51 $\pm$ \tiny{0.20} \\
        CFRNet$^*$ \cite{shalit2017estimating} & 2.72 $\pm$ \tiny{0.19} & 2.98 $\pm$ \tiny{0.22} & 1.48 $\pm$ \tiny{0.15} & 1.57 $\pm$ \tiny{0.17} & 4.18 $\pm$ \tiny{0.29} & 4.42 $\pm$ \tiny{0.31} & 2.72 $\pm$ \tiny{0.20} & 2.98 $\pm$ \tiny{0.23} & 3.98 $\pm$ \tiny{0.27} & 4.22 $\pm$ \tiny{0.30} & 2.32 $\pm$ \tiny{0.17} & 2.44 $\pm$ \tiny{0.19} \\
        DiffPO \cite{ma2024diffpo} & 2.03 $\pm$ \tiny{0.16} & 2.24 $\pm$ \tiny{0.14} & 1.25 $\pm$ \tiny{0.11} & 1.28 $\pm$ \tiny{0.09} & 3.85 $\pm$ \tiny{0.28} & 3.95 $\pm$ \tiny{0.29} & 2.18 $\pm$ \tiny{0.16} & 2.83 $\pm$ \tiny{0.15} & 3.64 $\pm$ \tiny{0.25} & 3.72 $\pm$ \tiny{0.26} & 1.89 $\pm$ \tiny{0.14} & 1.94 $\pm$ \tiny{0.13} \\
\midrule
\textbf{\methodshort (ours)} & \textbf{1.25 $\pm$ {\tiny 0.14}} & \textbf{1.26 $\pm$ {\tiny 0.11}} & \textbf{0.29 $\pm$ {\tiny 0.09}} & \textbf{0.30 $\pm$ {\tiny 0.06}} & \textbf{3.56 $\pm$ {\tiny 0.26}} & \textbf{3.49 $\pm$ {\tiny 0.25}} & \textbf{1.58 $\pm$ {\tiny 0.14}} & \textbf{1.60  $\pm$ {\tiny 0.12}} & \textbf{3.37 $\pm$ {\tiny 0.23}} & \textbf{3.25 $\pm$ {\tiny 0.24}} & \textbf{1.24 $\pm$ {\tiny 0.12}} & \textbf{1.30 $\pm$ {\tiny 0.11}} \\
\bottomrule
\multicolumn{13}{l}{Lower $=$ better ({best in bold}). $^*$ modified method to make it comparable. }
\end{tabular}%
}
\end{table*}


\begin{table*}[h]
\centering
\caption{Results showing in- \& out-of-sample empirical KL divergence (i.e., $\hat{D}_{\mathrm{KL}}^{\text{in}}$ and $\hat{D}_{\mathrm{KL}}^{\text{out}}$) for different intervention ($a = 0$ and $a = 1$) on ACIC, IST and MIMIC datasets. Reported: mean $\pm$ standard deviation over ten-fold train-test splits.}
\label{tab:results_kl}
\resizebox{0.95\textwidth}{!}{%
\begin{tabular}{l|rr|rr|rr|rr|rr|rr}
\toprule
{} & \multicolumn{4}{c|}{ACIC} & \multicolumn{4}{c|}{IST} & \multicolumn{4}{c}{MIMIC} \\
\cmidrule{2-5}\cmidrule{6-9}\cmidrule{10-13}
{} & \multicolumn{2}{c|}{$a = 0$} & \multicolumn{2}{c|}{$a = 1$} & \multicolumn{2}{c|}{$a = 0$} & \multicolumn{2}{c|}{$a = 1$} & \multicolumn{2}{c|}{$a = 0$} & \multicolumn{2}{c}{$a = 1$} \\
{} &  \multicolumn{1}{c}{$\hat{D}_{\mathrm{KL}}^\text{in}$} &  \multicolumn{1}{c|}{$\hat{D}_{\mathrm{KL}}^\text{out}$} &   \multicolumn{1}{c}{$\hat{D}_{\mathrm{KL}}^\text{in}$} &  \multicolumn{1}{c|}{$\hat{D}_{\mathrm{KL}}^\text{out}$} &  \multicolumn{1}{c}{$\hat{D}_{\mathrm{KL}}^\text{in}$} &  \multicolumn{1}{c|}{$\hat{D}_{\mathrm{KL}}^\text{out}$} &   \multicolumn{1}{c}{$\hat{D}_{\mathrm{KL}}^\text{in}$} &  \multicolumn{1}{c|}{$\hat{D}_{\mathrm{KL}}^\text{out}$} & \multicolumn{1}{c}{$\hat{D}_{\mathrm{KL}}^\text{in}$} &  \multicolumn{1}{c|}{$\hat{D}_{\mathrm{KL}}^\text{out}$} &   \multicolumn{1}{c}{$\hat{D}_{\mathrm{KL}}^\text{in}$} &  \multicolumn{1}{c}{$\hat{D}_{\mathrm{KL}}^\text{out}$}\\
\midrule
        S-Net$^*$ \cite{kunzel2019metalearners} & 2.67 $\pm$ \tiny{0.12} & 2.89 $\pm$ \tiny{0.15} & 2.36 $\pm$ \tiny{0.11} & 2.45 $\pm$ \tiny{0.14} & 5.12 $\pm$ \tiny{0.21} & 5.88 $\pm$ \tiny{0.28} & 4.43 $\pm$ \tiny{0.19} & 4.92 $\pm$ \tiny{0.25} & 4.56 $\pm$ \tiny{0.18} & 4.82 $\pm$ \tiny{0.24} & 3.89 $\pm$ \tiny{0.15} & 3.95 $\pm$ \tiny{0.20} \\
        T-Net$^*$ \cite{kunzel2019metalearners} & 2.28 $\pm$ \tiny{0.13} & 2.63 $\pm$ \tiny{0.16} & 2.05 $\pm$ \tiny{0.12} & 2.16 $\pm$ \tiny{0.15} & 5.01 $\pm$ \tiny{0.22} & 5.42 $\pm$ \tiny{0.29} & 4.27 $\pm$ \tiny{0.20} & 4.86 $\pm$ \tiny{0.26} & 4.20 $\pm$ \tiny{0.19} & 4.14 $\pm$ \tiny{0.22} & 3.46 $\pm$ \tiny{0.11} & 3.83 $\pm$ \tiny{0.12} \\
        GANITE$^*$ \cite{yoon2018ganite} & 3.90 $\pm$ \tiny{0.35} & 3.85 $\pm$ \tiny{0.39} & 3.68 $\pm$ \tiny{0.44} & 3.42 $\pm$ \tiny{0.47} & 7.25 $\pm$ \tiny{0.46} & 7.15 $\pm$ \tiny{0.45} & 5.92 $\pm$ \tiny{0.55} & 6.34 $\pm$ \tiny{0.52} & 6.65 $\pm$ \tiny{0.33} & 6.28 $\pm$ \tiny{0.31} & 4.95 $\pm$ \tiny{0.40} & 5.45 $\pm$ \tiny{0.47} \\
        TARNet$^*$ \cite{shalit2017estimating} & 2.42 $\pm$ \tiny{0.13} & 2.54 $\pm$ \tiny{0.17} & 2.18 $\pm$ \tiny{0.12} & 1.85 $\pm$ \tiny{0.15} & 4.88 $\pm$ \tiny{0.23} & 4.92 $\pm$ \tiny{0.31} & 4.21 $\pm$ \tiny{0.21} & 4.48 $\pm$ \tiny{0.27} & 3.92 $\pm$ \tiny{0.20} & 4.27 $\pm$ \tiny{0.26} & 3.28 $\pm$ \tiny{0.17} & 3.58 $\pm$ \tiny{0.23} \\
        CFRNet$^*$ \cite{shalit2017estimating} & 2.35 $\pm$ \tiny{0.12} & 2.47 $\pm$ \tiny{0.16} & 2.12 $\pm$ \tiny{0.11} & 1.80 $\pm$ \tiny{0.14} & 4.87 $\pm$ \tiny{0.22} & 4.98 $\pm$ \tiny{0.31} & 4.15 $\pm$ \tiny{0.21} & 4.33 $\pm$ \tiny{0.27} & 3.85 $\pm$ \tiny{0.19} & 4.15 $\pm$ \tiny{0.26} & 3.14 $\pm$ \tiny{0.16} & 3.42 $\pm$ \tiny{0.22} \\
        DiffPO \cite{ma2024diffpo} & 1.97 $\pm$ \tiny{0.07} & 1.85 $\pm$ \tiny{0.10} & 0.69 $\pm$ \tiny{0.09} & 1.38 $\pm$ \tiny{0.08} & 4.21 $\pm$ \tiny{0.15} & 4.25 $\pm$ \tiny{0.22} & 3.41 $\pm$ \tiny{0.13} & 3.57 $\pm$ \tiny{0.18} & 3.35 $\pm$ \tiny{0.12} & 3.55 $\pm$ \tiny{0.18} & 1.85 $\pm$ \tiny{0.10} & 2.02 $\pm$ \tiny{0.15} \\
\midrule
\textbf{\methodshort (ours)} & \textbf{1.08 $\pm$ {\tiny 0.05}} & \textbf{1.22 $\pm$ {\tiny 0.07}} & \textbf{0.11 $\pm$ {\tiny 0.08}} & \textbf{0.09 $\pm$ {\tiny 0.05}} & \textbf{3.28 $\pm$ {\tiny 0.12}} & \textbf{3.76 $\pm$ {\tiny 0.19}} & \textbf{2.92 $\pm$ {\tiny 0.12}} & \textbf{2.85 $\pm$ {\tiny 0.16}} & \textbf{2.84 $\pm$ {\tiny 0.10}} & \textbf{2.99 $\pm$ {\tiny 0.15}} & \textbf{1.44 $\pm$ {\tiny 0.07}} & \textbf{1.46 $\pm$ {\tiny 0.08}} \\
\bottomrule
\multicolumn{13}{l}{Lower $=$ better ({best in bold}). $^*$ modified method to make it comparable. }
\end{tabular}%
}
\end{table*}

\subsection{Baselines} 

Due to the novelty of our method, there are \underline{no} existing methods tailored for this task, namely, learning the joint interventional distribution (see Section~\ref{sec:related_work}). As a remedy, we adapt existing methods from two streams to our task. 

\emph{Benchmarks for distributions:} We use methods that learn the interventional distribution of CAPOs. As such, these benchmarks preserve the causal structure and thus allow for a fair comparison. Here, \textbf{DiffPO} \cite{ma2024diffpo} is our main baseline. To the best of our knowledge, DiffPO is the only baseline method that is designed for learning the distribution of CAPOs, but only in single-outcome settings. Another limitation is that this method is constrained to continuous outcomes and is thus \emph{not} able to deal with mixed-type outcomes. As a remedy, we adapted the method by approximating the joint interventional distribution by the product of marginal distributions.  

\emph{Benchmarks for CAPOs:} We also use benchmarks that are designed for computing point estimates of CAPOs but which we adapt to handle distributions. (1)~\textbf{GANITE} \cite{yoon2018ganite}: uses a generative adversarial network to generate potential outcomes, yet where the method was not designed for learning the distribution. We adapt GANITE to our task by removing the second stage, so that we directly sample potential outcomes from the learned distributions in the first stage. Then, we approximate the joint distribution by the product of the marginals learned by separate models. (2)~\textbf{S-Net} \cite{kunzel2019metalearners}: the S-learner is a model-agnostic learner that trains a single regression model by concatenating the covariate and the treatment as input, which we instantiate with neural networks; (2)~\textbf{T-Net} \cite{kunzel2019metalearners}: the T-learner is a model-agnostic learner that trains separate regression models for treated and control groups, which we instantiate with neural networks; (4)~\textbf{TARNet} \cite{shalit2017estimating}: using representation learning to extract features of covariates and train separate branches for treated and control groups with regularization. (5)~\textbf{CFRNet} \cite{shalit2017estimating}: is based on representation learning used in variants of balancing with TARNet. Of note,  methods (2)--(5) are only able to give point estimates. Therefore, we follow prior work \cite{hess2023bayesian, ma2024diffpo} to extend them with Monte Carlo (MC) dropout to make these methods comparable to our method. As these methods initially targeted for a single outcome, we again approximate the joint interventional distribution by the product of marginal distributions. Further implementation details are in Appendix~\ref{app:baseline_implementation}.

\subsection{Datasets}

\subsubsection{Datasets for benchmarking causal inference}

We use \textbf{ACIC2018} \cite{macdorman1998infant}, which is a benchmarking dataset specifically tailored to causal machine learning. 
Covariates of ACIC2018 are derived from the linked birth and infant death data. Detailed introduction of the datasets can be found in Appendix~\ref{app:all_dataset}. Due to the fundamental problem of causal inference, the counterfactual outcomes can not be observed in real-world data. We thus adapt the dataset to our task by using the real-world covariates and generating outcomes following prior literature (e.g., \cite{kennedy2019estimating, shalit2017estimating, melnychuk2023normalizing, ma2024diffpo}). Further details are provided in the Appendix~\ref{app:syn-data}.

\subsubsection{Medical datasets}

We further use the following medical datasets:

\begin{itemize}[leftmargin=15pt]
\item \textbf{IST dataset:} The International Stroke Trial (IST) \cite{sandercock2011international} is one of the largest randomized controlled trials in acute stroke treatment. The dataset comprises 19,435 patients. The treatment assignments include trial aspirin allocation and trial heparin allocation. The outcomes are the probability of new strokes due to blocked vessels and death or dependency at 6 months. Recurrent strokes may result in death and requiring assistance for daily activities. Covariates include age, sex, presence of atrial fibrillation, systolic blood pressure, infarct visibility on CT, prior heparin use, prior aspirin use, and recorded deficits of different body parts. 
\vspace{0.2cm}

\item \textbf{MIMIC-III dataset:} MIMIC-III \cite{johnson2016mimic} is a large, single-center database comprising information relating to patients admitted to critical care units at a large tertiary care hospital. MIMIC-III contains 38,597 distinct adult patients. The treatment is mechanical ventilation. The outcomes are the length of hospital stays and long-term mortality outcome (90-day mortality). Survivors typically had shorter hospital stays. Confounders include demographic variables and clinical variables (e.g., glucose, hematocrit, creatinine, sodium, blood urea nitrogen, hemoglobin, heart rate, mean blood pressure, platelets, respiratory rate, bicarbonate, red blood cell count, and anion gap).  
\end{itemize}

\subsection{Results}
The results of evaluating the performance in learning the distributions are in Table~\ref{tab:results_wass} and Table~\ref{tab:results_kl}. We find that our method gives the lowest empirical Wasserstein distance and empirical KL divergence out of all methods, which is desirable. Hence, the experiments show that \emph{our method outperforms the baselines by a clear margin}.

\subsection{Flexibility to handle other causal quantities}

A key strength of our method is its flexibility. It can be used not only to model multi-outcome distributions, but also to estimate other causal quantities. For example, it also supports point estimation tasks. In our evaluation, we compute the average conditional average treatment effect (CATE) across $k$ outcomes. We report the precision of estimating the heterogeneous effects (PEHE) criterion \cite{hill2011bayesian, curth2021nonparametric} in Table~\ref{tab:results_cate}. The experiments demonstrate that our method achieves state-of-the-art performance in CATE estimation for the multiple outcomes task.

\begin{table}[h]
\centering
\caption{Results showing the averaged CATE evaluation performance on IST and MIMIC datasets.}
\label{tab:results_cate}
\resizebox{0.23 \textwidth}{!}{%
\begin{tabular}{l|cc}
\toprule
 & IST & MIMIC \\
\midrule
S-Net \cite{kunzel2019metalearners}& 1.23 $\pm$ {\tiny 0.05} & 0.96 $\pm$ {\tiny 0.05} \\
T-Net \cite{kunzel2019metalearners} & 1.21 $\pm$ {\tiny 0.07} & 0.93 $\pm$ {\tiny 0.06} \\
GANITE \cite{yoon2018ganite} & 2.64 $\pm$ {\tiny 0.34} & 1.55 $\pm$ {\tiny 0.26} \\
TARNet \cite{shalit2017estimating} & 1.19 $\pm$ {\tiny 0.04} & 0.90 $\pm$ {\tiny 0.04} \\
CFRNet \cite{shalit2017estimating} & 1.18 $\pm$ {\tiny 0.04} & 0.89 $\pm$ {\tiny 0.04} \\
DiffPO \cite{ma2024diffpo} & 1.20 $\pm$ {\tiny 0.15} & 0.92 $\pm$ {\tiny 0.12} \\
\midrule
\textbf{DIME (ours)} & \textbf{1.15} $\pm$ {\tiny 0.10} & \textbf{0.87} $\pm$ {\tiny 0.08} \\
\bottomrule
\multicolumn{3}{l}{Lower = better (best in bold).}
\end{tabular}%
}
\end{table}

\subsection{Visualization}

\begin{figure}[htbp]
\begin{center}
\centerline{\includegraphics[width=0.9\columnwidth]{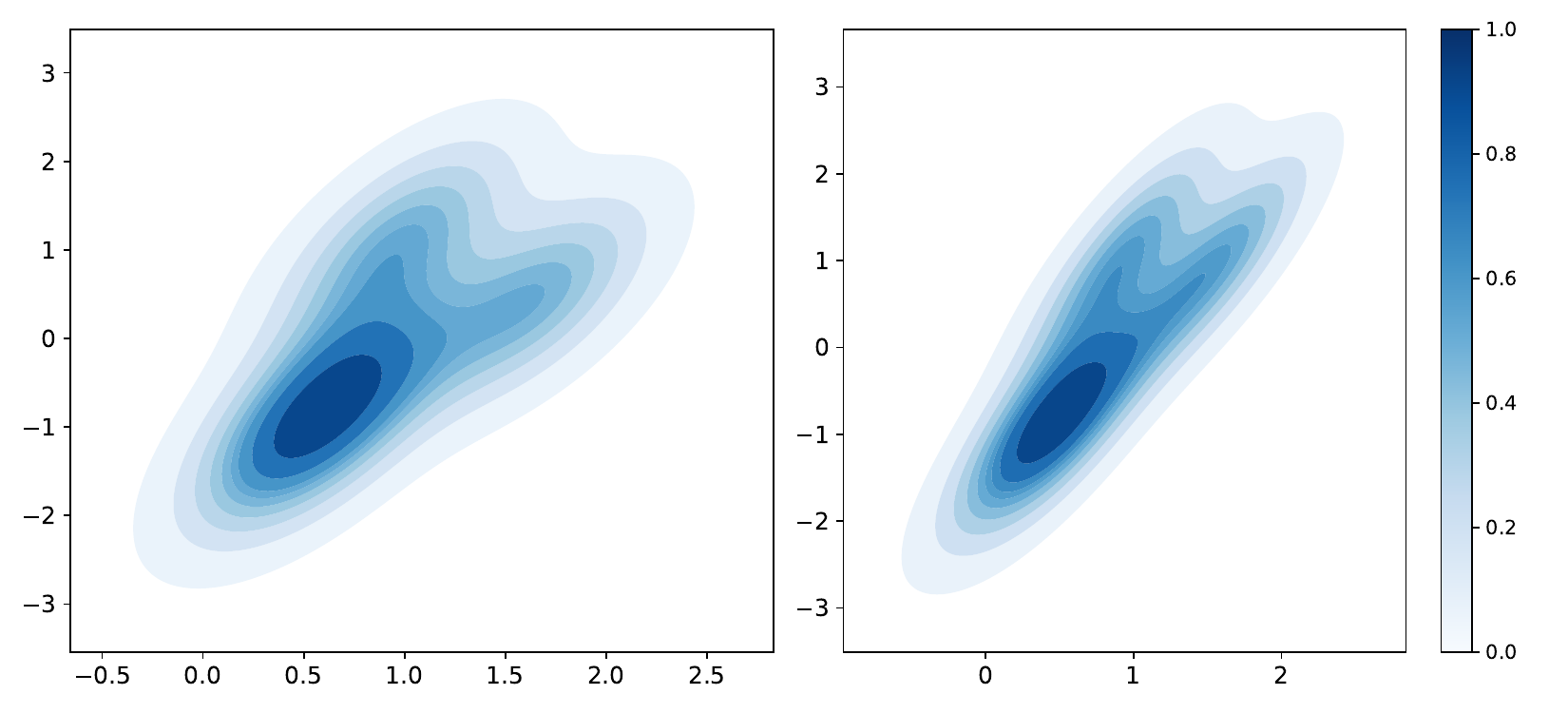}}
\caption{Kernel density estimation plot of outcomes on MIMIC dataset. \emph{Left:} our model prediction; \emph{Right:} the ground-truth distribution.}
\label{fig:mimic_0}
\end{center}
\end{figure}

It is crucial in medical practice \cite{heckman1997making,zampieri2021using, banerji2023clinical, kneib2023rage} to understand the expected probability of whether a treatment is beneficial and thus to better assess the reliability of the estimates. We present the visualization of learned distributions for a real-world dataset from medicine. Learning such distributions is inherently challenging, particularly in observational datasets where only a single sample per individual $X = x$ is available for training the model. Consequently, our method must predict the entire joint distribution $p(Y^1(a), Y^2(a) \mid X=x)$ from a single observation of outcomes ($y^1$ and $y^2$ given $x$) by inferring the underlying relationships. As illustrated in Fig.~\ref{fig:mimic_0}, our method successfully reconstructs the full distribution that closely approximates the ground truth. This demonstrates its capability to capture comprehensive information about the outcome distributions, providing valuable insights for medical decision-making.

\vspace{-0.5cm}

\section{Conclusion}

Our method is carefully designed for learning the joint interventional distribution of multiple outcomes in medical practice. It paves the way for reliable, uncertainty-aware decision-making in personalized medicine.

\begin{acks}
This work has been supported by the German Federal Ministry of Education and Research (Grant: 01IS24082).
\end{acks}

\clearpage
\bibliographystyle{ACM-Reference-Format-no-doi}
\bibliography{kdd-references}


\begin{thebibliography}{56}


\ifx \showCODEN    \undefined \def \showCODEN     #1{\unskip}     \fi
\ifx \showDOI      \undefined \def \showDOI       #1{#1}\fi
\ifx \showISBNx    \undefined \def \showISBNx     #1{\unskip}     \fi
\ifx \showISBNxiii \undefined \def \showISBNxiii  #1{\unskip}     \fi
\ifx \showISSN     \undefined \def \showISSN      #1{\unskip}     \fi
\ifx \showLCCN     \undefined \def \showLCCN      #1{\unskip}     \fi
\ifx \shownote     \undefined \def \shownote      #1{#1}          \fi
\ifx \showarticletitle \undefined \def \showarticletitle #1{#1}   \fi
\ifx \showURL      \undefined \def \showURL       {\relax}        \fi
\providecommand\bibfield[2]{#2}
\providecommand\bibinfo[2]{#2}
\providecommand\natexlab[1]{#1}
\providecommand\showeprint[2][]{arXiv:#2}

\bibitem[\protect\citeauthoryear{Banerji, Chakraborti, Harbron, and MacArthur}{Banerji et~al\mbox{.}}{2023}]%
        {banerji2023clinical}
\bibfield{author}{\bibinfo{person}{Christopher~RS Banerji}, \bibinfo{person}{Tapabrata Chakraborti}, \bibinfo{person}{Chris Harbron}, {and} \bibinfo{person}{Ben~D MacArthur}.} \bibinfo{year}{2023}\natexlab{}.
\newblock \showarticletitle{Clinical AI tools must convey predictive uncertainty for each individual patient}.
\newblock \bibinfo{journal}{\emph{Nature Medicine}} \bibinfo{volume}{29}, \bibinfo{number}{12} (\bibinfo{year}{2023}), \bibinfo{pages}{2996--2998}.
\newblock


\bibitem[\protect\citeauthoryear{Byambadalai, Oka, and Yasui}{Byambadalai et~al\mbox{.}}{2024}]%
        {byambadalai2024estimating}
\bibfield{author}{\bibinfo{person}{Undral Byambadalai}, \bibinfo{person}{Tatsushi Oka}, {and} \bibinfo{person}{Shota Yasui}.} \bibinfo{year}{2024}\natexlab{}.
\newblock \showarticletitle{Estimating Distributional Treatment Effects in Randomized Experiments: Machine Learning for Variance Reduction}. In \bibinfo{booktitle}{\emph{International Conference on Machine Learning}}.
\newblock


\bibitem[\protect\citeauthoryear{Chipman, George, and McCulloch}{Chipman et~al\mbox{.}}{2010}]%
        {chipman2010bart}
\bibfield{author}{\bibinfo{person}{Hugh~A. Chipman}, \bibinfo{person}{Edward~I. George}, {and} \bibinfo{person}{Robert~E. McCulloch}.} \bibinfo{year}{2010}\natexlab{}.
\newblock \showarticletitle{{BART}: Bayesian additive regression trees}.
\newblock \bibinfo{journal}{\emph{The Annals of Applied Statistics}} \bibinfo{volume}{4}, \bibinfo{number}{1} (\bibinfo{date}{March} \bibinfo{year}{2010}), \bibinfo{pages}{266--298}.
\newblock


\bibitem[\protect\citeauthoryear{Curth and van~der Schaar}{Curth and van~der Schaar}{2021a}]%
        {curth2021nonparametric}
\bibfield{author}{\bibinfo{person}{Alicia Curth} {and} \bibinfo{person}{Mihaela van~der Schaar}.} \bibinfo{year}{2021}\natexlab{a}.
\newblock \showarticletitle{Nonparametric estimation of heterogeneous treatment effects: From theory to learning algorithms}. In \bibinfo{booktitle}{\emph{International Conference on Artificial Intelligence and Statistics}}.
\newblock


\bibitem[\protect\citeauthoryear{Curth and van~der Schaar}{Curth and van~der Schaar}{2021b}]%
        {curth2021inductive}
\bibfield{author}{\bibinfo{person}{Alicia Curth} {and} \bibinfo{person}{Mihaela van~der Schaar}.} \bibinfo{year}{2021}\natexlab{b}.
\newblock \showarticletitle{On inductive biases for heterogeneous treatment effect estimation}.
\newblock \bibinfo{journal}{\emph{Advances in Neural Information Processing Systems}} (\bibinfo{year}{2021}).
\newblock


\bibitem[\protect\citeauthoryear{Doutreligne, Struja, Abecassis, Morgand, Celi, and Varoquaux}{Doutreligne et~al\mbox{.}}{2025}]%
        {doutreligne2025step}
\bibfield{author}{\bibinfo{person}{Matthieu Doutreligne}, \bibinfo{person}{Tristan Struja}, \bibinfo{person}{Judith Abecassis}, \bibinfo{person}{Claire Morgand}, \bibinfo{person}{Leo~Anthony Celi}, {and} \bibinfo{person}{Ga{\"e}l Varoquaux}.} \bibinfo{year}{2025}\natexlab{}.
\newblock \showarticletitle{Step-by-step causal analysis of EHRs to ground decision-making}.
\newblock \bibinfo{journal}{\emph{PLOS Digital Health}} \bibinfo{volume}{4}, \bibinfo{number}{2} (\bibinfo{year}{2025}), \bibinfo{pages}{e0000721}.
\newblock


\bibitem[\protect\citeauthoryear{Feliu, Heredia-Soto, Giron{\'e}s, Jim{\'e}nez-Munarriz, Salda{\~n}a, Guill{\'e}n-Ponce, and Molina-Garrido}{Feliu et~al\mbox{.}}{2020}]%
        {feliu2020management}
\bibfield{author}{\bibinfo{person}{J Feliu}, \bibinfo{person}{V Heredia-Soto}, \bibinfo{person}{R Giron{\'e}s}, \bibinfo{person}{B Jim{\'e}nez-Munarriz}, \bibinfo{person}{J Salda{\~n}a}, \bibinfo{person}{C Guill{\'e}n-Ponce}, {and} \bibinfo{person}{MJ Molina-Garrido}.} \bibinfo{year}{2020}\natexlab{}.
\newblock \showarticletitle{Management of the toxicity of chemotherapy and targeted therapies in elderly cancer patients}.
\newblock \bibinfo{journal}{\emph{Clinical and Translational Oncology}}  \bibinfo{volume}{22} (\bibinfo{year}{2020}), \bibinfo{pages}{457--467}.
\newblock


\bibitem[\protect\citeauthoryear{Feuerriegel, Frauen, Melnychuk, Schweisthal, Hess, Curth, Bauer, Kilbertus, Kohane, and van~der Schaar}{Feuerriegel et~al\mbox{.}}{2024}]%
        {Feuerriegel2024}
\bibfield{author}{\bibinfo{person}{Stefan Feuerriegel}, \bibinfo{person}{Dennis Frauen}, \bibinfo{person}{Valentyn Melnychuk}, \bibinfo{person}{Jonas Schweisthal}, \bibinfo{person}{Konstantin Hess}, \bibinfo{person}{Alicia Curth}, \bibinfo{person}{Stefan Bauer}, \bibinfo{person}{Niki Kilbertus}, \bibinfo{person}{Isaac~S Kohane}, {and} \bibinfo{person}{Mihaela van~der Schaar}.} \bibinfo{year}{2024}\natexlab{}.
\newblock \showarticletitle{Causal machine learning for predicting treatment outcomes}.
\newblock \bibinfo{journal}{\emph{Nature Medicine}} \bibinfo{volume}{30}, \bibinfo{number}{4} (\bibinfo{year}{2024}), \bibinfo{pages}{958--968}.
\newblock


\bibitem[\protect\citeauthoryear{Frei~III and Canellos}{Frei~III and Canellos}{1980}]%
        {frei1980dose}
\bibfield{author}{\bibinfo{person}{Emil Frei~III} {and} \bibinfo{person}{George~P Canellos}.} \bibinfo{year}{1980}\natexlab{}.
\newblock \showarticletitle{Dose: a critical factor in cancer chemotherapy}.
\newblock \bibinfo{journal}{\emph{The American Journal of Medicine}} \bibinfo{volume}{69}, \bibinfo{number}{4} (\bibinfo{year}{1980}), \bibinfo{pages}{585--594}.
\newblock


\bibitem[\protect\citeauthoryear{Heckman, Smith, and Clements}{Heckman et~al\mbox{.}}{1997}]%
        {heckman1997making}
\bibfield{author}{\bibinfo{person}{James~J. Heckman}, \bibinfo{person}{Jeffrey Smith}, {and} \bibinfo{person}{Nancy Clements}.} \bibinfo{year}{1997}\natexlab{}.
\newblock \showarticletitle{Making the most out of programme evaluations and social experiments: Accounting for heterogeneity in programme impacts}.
\newblock \bibinfo{journal}{\emph{The Review of Economic Studies}} \bibinfo{volume}{64}, \bibinfo{number}{4} (\bibinfo{year}{1997}), \bibinfo{pages}{487--535}.
\newblock


\bibitem[\protect\citeauthoryear{Hess, Melnychuk, Frauen, and Feuerriegel}{Hess et~al\mbox{.}}{2023}]%
        {hess2023bayesian}
\bibfield{author}{\bibinfo{person}{Konstantin Hess}, \bibinfo{person}{Valentyn Melnychuk}, \bibinfo{person}{Dennis Frauen}, {and} \bibinfo{person}{Stefan Feuerriegel}.} \bibinfo{year}{2023}\natexlab{}.
\newblock \showarticletitle{Bayesian neural controlled differential equations for treatment effect estimation}. In \bibinfo{booktitle}{\emph{International Conference on Learning Representations}}.
\newblock


\bibitem[\protect\citeauthoryear{Hill}{Hill}{2011}]%
        {hill2011bayesian}
\bibfield{author}{\bibinfo{person}{Jennifer~L. Hill}.} \bibinfo{year}{2011}\natexlab{}.
\newblock \showarticletitle{Bayesian nonparametric modeling for causal inference}.
\newblock \bibinfo{journal}{\emph{Journal of Computational and Graphical Statistics}} \bibinfo{volume}{20}, \bibinfo{number}{1} (\bibinfo{year}{2011}), \bibinfo{pages}{217--240}.
\newblock


\bibitem[\protect\citeauthoryear{Ho, Jain, and Abbeel}{Ho et~al\mbox{.}}{2020}]%
        {ho2020denoising}
\bibfield{author}{\bibinfo{person}{Jonathan Ho}, \bibinfo{person}{Ajay Jain}, {and} \bibinfo{person}{Pieter Abbeel}.} \bibinfo{year}{2020}\natexlab{}.
\newblock \showarticletitle{Denoising diffusion probabilistic models}. In \bibinfo{booktitle}{\emph{Advances in Neural Information Processing Systems}}.
\newblock


\bibitem[\protect\citeauthoryear{Holland}{Holland}{1986}]%
        {holland1986statistics}
\bibfield{author}{\bibinfo{person}{Paul~W Holland}.} \bibinfo{year}{1986}\natexlab{}.
\newblock \showarticletitle{Statistics and causal inference}.
\newblock \bibinfo{journal}{\emph{Journal of the American Statistical Association}} \bibinfo{volume}{81}, \bibinfo{number}{396} (\bibinfo{year}{1986}), \bibinfo{pages}{945--960}.
\newblock


\bibitem[\protect\citeauthoryear{Johansson, Kallus, Shalit, and Sontag}{Johansson et~al\mbox{.}}{2018}]%
        {johansson2018learning}
\bibfield{author}{\bibinfo{person}{Fredrik~D. Johansson}, \bibinfo{person}{Nathan Kallus}, \bibinfo{person}{Uri Shalit}, {and} \bibinfo{person}{David Sontag}.} \bibinfo{year}{2018}\natexlab{}.
\newblock \showarticletitle{Learning weighted representations for generalization across designs}.
\newblock \bibinfo{journal}{\emph{arXiv preprint}} (\bibinfo{year}{2018}).
\newblock


\bibitem[\protect\citeauthoryear{Johansson, Shalit, and Sontag}{Johansson et~al\mbox{.}}{2016}]%
        {johansson2016learning}
\bibfield{author}{\bibinfo{person}{Fredrik~D. Johansson}, \bibinfo{person}{Uri Shalit}, {and} \bibinfo{person}{David Sontag}.} \bibinfo{year}{2016}\natexlab{}.
\newblock \showarticletitle{Learning representations for counterfactual inference}. In \bibinfo{booktitle}{\emph{International Conference on Machine Learning}}.
\newblock


\bibitem[\protect\citeauthoryear{Johnson, Pollard, Shen, Lehman, Feng, Ghassemi, Moody, Szolovits, Celi, and Mark}{Johnson et~al\mbox{.}}{2016}]%
        {johnson2016mimic}
\bibfield{author}{\bibinfo{person}{Alistair~EW Johnson}, \bibinfo{person}{Tom~J Pollard}, \bibinfo{person}{Lu Shen}, \bibinfo{person}{Li-wei~H Lehman}, \bibinfo{person}{Mengling Feng}, \bibinfo{person}{Mohammad Ghassemi}, \bibinfo{person}{Benjamin Moody}, \bibinfo{person}{Peter Szolovits}, \bibinfo{person}{Leo~Anthony Celi}, {and} \bibinfo{person}{Roger~G Mark}.} \bibinfo{year}{2016}\natexlab{}.
\newblock \showarticletitle{MIMIC-III, a freely accessible critical care database}.
\newblock \bibinfo{journal}{\emph{Scientific Data}}  \bibinfo{volume}{3} (\bibinfo{year}{2016}), \bibinfo{pages}{160035}.
\newblock


\bibitem[\protect\citeauthoryear{Katout, Zhu, Rutsky, Shah, Brook, Zhong, and Rajagopalan}{Katout et~al\mbox{.}}{2014}]%
        {katout2014effect}
\bibfield{author}{\bibinfo{person}{Mohammad Katout}, \bibinfo{person}{Hong Zhu}, \bibinfo{person}{Jessica Rutsky}, \bibinfo{person}{Parthy Shah}, \bibinfo{person}{Robert~D Brook}, \bibinfo{person}{Jixin Zhong}, {and} \bibinfo{person}{Sanjay Rajagopalan}.} \bibinfo{year}{2014}\natexlab{}.
\newblock \showarticletitle{Effect of GLP-1 mimetics on blood pressure and relationship to weight loss and glycemia lowering: results of a systematic meta-analysis and meta-regression}.
\newblock \bibinfo{journal}{\emph{American Journal of Hypertension}} \bibinfo{volume}{27}, \bibinfo{number}{1} (\bibinfo{year}{2014}), \bibinfo{pages}{130--139}.
\newblock


\bibitem[\protect\citeauthoryear{Kennedy}{Kennedy}{2023}]%
        {kennedy2023towards}
\bibfield{author}{\bibinfo{person}{Edward~H. Kennedy}.} \bibinfo{year}{2023}\natexlab{}.
\newblock \showarticletitle{Towards optimal doubly robust estimation of heterogeneous causal effects}.
\newblock \bibinfo{journal}{\emph{Electronic Journal of Statistics}} \bibinfo{volume}{17}, \bibinfo{number}{2} (\bibinfo{year}{2023}), \bibinfo{pages}{3008--3049}.
\newblock


\bibitem[\protect\citeauthoryear{Kennedy, Kangovi, and Mitra}{Kennedy et~al\mbox{.}}{2019}]%
        {kennedy2019estimating}
\bibfield{author}{\bibinfo{person}{Edward~H Kennedy}, \bibinfo{person}{Shreya Kangovi}, {and} \bibinfo{person}{Nandita Mitra}.} \bibinfo{year}{2019}\natexlab{}.
\newblock \showarticletitle{Estimating scaled treatment effects with multiple outcomes}.
\newblock \bibinfo{journal}{\emph{Statistical Methods in Medical Research}} \bibinfo{volume}{28}, \bibinfo{number}{4} (\bibinfo{year}{2019}), \bibinfo{pages}{1094--1104}.
\newblock


\bibitem[\protect\citeauthoryear{Kern, Fischer-Abaigar, Schweisthal, Frauen, Ghani, Feuerriegel, van~der Schaar, and Kreuter}{Kern et~al\mbox{.}}{2025}]%
        {Kern2025}
\bibfield{author}{\bibinfo{person}{Christoph Kern}, \bibinfo{person}{Unai Fischer-Abaigar}, \bibinfo{person}{Jonas Schweisthal}, \bibinfo{person}{Dennis Frauen}, \bibinfo{person}{Rayid Ghani}, \bibinfo{person}{Stefan Feuerriegel}, \bibinfo{person}{Mihaela van~der Schaar}, {and} \bibinfo{person}{Frauke Kreuter}.} \bibinfo{year}{2025}\natexlab{}.
\newblock \showarticletitle{Algorithms for reliable decision-making need causal reasoning}.
\newblock \bibinfo{journal}{\emph{Nature Computational Science}} (\bibinfo{year}{2025}).
\newblock


\bibitem[\protect\citeauthoryear{Kneib, Silbersdorff, and S{\"a}fken}{Kneib et~al\mbox{.}}{2023}]%
        {kneib2023rage}
\bibfield{author}{\bibinfo{person}{Thomas Kneib}, \bibinfo{person}{Alexander Silbersdorff}, {and} \bibinfo{person}{Benjamin S{\"a}fken}.} \bibinfo{year}{2023}\natexlab{}.
\newblock \showarticletitle{Rage against the mean: a review of distributional regression approaches}.
\newblock \bibinfo{journal}{\emph{Econometrics and Statistics}}  \bibinfo{volume}{26} (\bibinfo{year}{2023}), \bibinfo{pages}{99--123}.
\newblock


\bibitem[\protect\citeauthoryear{K{\"u}nzel, Sekhon, Bickel, and Yu}{K{\"u}nzel et~al\mbox{.}}{2019}]%
        {kunzel2019metalearners}
\bibfield{author}{\bibinfo{person}{S{\"o}ren~R. K{\"u}nzel}, \bibinfo{person}{Jasjeet~S. Sekhon}, \bibinfo{person}{Peter~J. Bickel}, {and} \bibinfo{person}{Bin Yu}.} \bibinfo{year}{2019}\natexlab{}.
\newblock \showarticletitle{Metalearners for estimating heterogeneous treatment effects using machine learning}.
\newblock \bibinfo{journal}{\emph{Proceedings of the National Academy of Sciences}} \bibinfo{volume}{116}, \bibinfo{number}{10} (\bibinfo{year}{2019}), \bibinfo{pages}{4156--4165}.
\newblock


\bibitem[\protect\citeauthoryear{Lee, Kim, and Park}{Lee et~al\mbox{.}}{2023}]%
        {lee2023codi}
\bibfield{author}{\bibinfo{person}{Chaejeong Lee}, \bibinfo{person}{Jayoung Kim}, {and} \bibinfo{person}{Noseong Park}.} \bibinfo{year}{2023}\natexlab{}.
\newblock \showarticletitle{CoDi: Co-evolving contrastive diffusion models for mixed-type tabular synthesis}. In \bibinfo{booktitle}{\emph{International Conference on Machine Learning}}.
\newblock


\bibitem[\protect\citeauthoryear{Lin, Ryan, Sammel, Zhang, Padungtod, and Xu}{Lin et~al\mbox{.}}{2000}]%
        {lin2000scaled}
\bibfield{author}{\bibinfo{person}{Xihong Lin}, \bibinfo{person}{Louise Ryan}, \bibinfo{person}{Mary Sammel}, \bibinfo{person}{Daowen Zhang}, \bibinfo{person}{Chantana Padungtod}, {and} \bibinfo{person}{Xiping Xu}.} \bibinfo{year}{2000}\natexlab{}.
\newblock \showarticletitle{A scaled linear mixed model for multiple outcomes}.
\newblock \bibinfo{journal}{\emph{Biometrics}} \bibinfo{volume}{56}, \bibinfo{number}{2} (\bibinfo{year}{2000}), \bibinfo{pages}{593--601}.
\newblock


\bibitem[\protect\citeauthoryear{Ma, Melnychuk, Schweisthal, and Feuerriegel}{Ma et~al\mbox{.}}{2024}]%
        {ma2024diffpo}
\bibfield{author}{\bibinfo{person}{Yuchen Ma}, \bibinfo{person}{Valentyn Melnychuk}, \bibinfo{person}{Jonas Schweisthal}, {and} \bibinfo{person}{Stefan Feuerriegel}.} \bibinfo{year}{2024}\natexlab{}.
\newblock \showarticletitle{DiffPO: A causal diffusion model for learning distributions of potential outcomes}. In \bibinfo{booktitle}{\emph{Advances in Neural Information Processing Systems}}.
\newblock


\bibitem[\protect\citeauthoryear{Maag, Feuerriegel, Kraus, Saar-Tsechansky, and Z{\"u}ger}{Maag et~al\mbox{.}}{2021}]%
        {maag2021modeling}
\bibfield{author}{\bibinfo{person}{Basil Maag}, \bibinfo{person}{Stefan Feuerriegel}, \bibinfo{person}{Mathias Kraus}, \bibinfo{person}{Maytal Saar-Tsechansky}, {and} \bibinfo{person}{Thomas Z{\"u}ger}.} \bibinfo{year}{2021}\natexlab{}.
\newblock \showarticletitle{Modeling longitudinal dynamics of comorbidities}. In \bibinfo{booktitle}{\emph{Conference on Health, Inference, and Learning}}.
\newblock


\bibitem[\protect\citeauthoryear{MacDorman and Atkinson}{MacDorman and Atkinson}{1998}]%
        {macdorman1998infant}
\bibfield{author}{\bibinfo{person}{M.F. MacDorman} {and} \bibinfo{person}{J.O. Atkinson}.} \bibinfo{year}{1998}\natexlab{}.
\newblock \showarticletitle{Infant mortality statistics from the linked birth/infant death data set--1995 period data}.
\newblock \bibinfo{journal}{\emph{Monthly Vital Statistics Report}} \bibinfo{volume}{46}, \bibinfo{number}{6 Suppl 2} (\bibinfo{year}{1998}), \bibinfo{pages}{1--22}.
\newblock


\bibitem[\protect\citeauthoryear{Melnychuk, Feuerriegel, and {van der Schaar}}{Melnychuk et~al\mbox{.}}{2024}]%
        {melnychuk2024quantifying}
\bibfield{author}{\bibinfo{person}{Valentyn Melnychuk}, \bibinfo{person}{Stefan Feuerriegel}, {and} \bibinfo{person}{Mihaela {van der Schaar}}.} \bibinfo{year}{2024}\natexlab{}.
\newblock \showarticletitle{Quantifying aleatoric uncertainty of the treatment effect: a novel orthogonal learner}. In \bibinfo{booktitle}{\emph{Advances in Neural Information Processing Systems}}.
\newblock


\bibitem[\protect\citeauthoryear{Melnychuk, Frauen, and Feuerriegel}{Melnychuk et~al\mbox{.}}{2023}]%
        {melnychuk2023normalizing}
\bibfield{author}{\bibinfo{person}{Valentyn Melnychuk}, \bibinfo{person}{Dennis Frauen}, {and} \bibinfo{person}{Stefan Feuerriegel}.} \bibinfo{year}{2023}\natexlab{}.
\newblock \showarticletitle{Normalizing flows for interventional density estimation}. In \bibinfo{booktitle}{\emph{International Conference on Machine Learning}}.
\newblock


\bibitem[\protect\citeauthoryear{Naumzik, Feuerriegel, and Nielsen}{Naumzik et~al\mbox{.}}{2023}]%
        {naumzik2023data}
\bibfield{author}{\bibinfo{person}{Christof Naumzik}, \bibinfo{person}{Stefan Feuerriegel}, {and} \bibinfo{person}{Anne~Molgaard Nielsen}.} \bibinfo{year}{2023}\natexlab{}.
\newblock \showarticletitle{Data-driven dynamic treatment planning for chronic diseases}.
\newblock \bibinfo{journal}{\emph{European Journal of Operational Research}} (\bibinfo{year}{2023}).
\newblock


\bibitem[\protect\citeauthoryear{Naumzik, Kongsted, Vach, and Feuerriegel}{Naumzik et~al\mbox{.}}{2024}]%
        {naumzik2024data}
\bibfield{author}{\bibinfo{person}{Christof Naumzik}, \bibinfo{person}{Alice Kongsted}, \bibinfo{person}{Werner Vach}, {and} \bibinfo{person}{Stefan Feuerriegel}.} \bibinfo{year}{2024}\natexlab{}.
\newblock \showarticletitle{Data-driven subgrouping of patient trajectories with chronic diseases: Evidence from low back pain}. In \bibinfo{booktitle}{\emph{Proceedings of the Conference on Health, Inference, and Learning (CHIL)}}.
\newblock


\bibitem[\protect\citeauthoryear{Nie and Wager}{Nie and Wager}{2021}]%
        {nie2021quasi}
\bibfield{author}{\bibinfo{person}{Xinkun Nie} {and} \bibinfo{person}{Stefan Wager}.} \bibinfo{year}{2021}\natexlab{}.
\newblock \showarticletitle{Quasi-oracle estimation of heterogeneous treatment effects}.
\newblock \bibinfo{journal}{\emph{Biometrika}}  \bibinfo{volume}{108} (\bibinfo{year}{2021}), \bibinfo{pages}{299--319}.
\newblock


\bibitem[\protect\citeauthoryear{Page, O’Bryant, Cheng, Dow, Ky, Stein, Spencer, Trupp, and Lindenfeld}{Page et~al\mbox{.}}{2016}]%
        {page2016drugs}
\bibfield{author}{\bibinfo{person}{Robert~L Page}, \bibinfo{person}{Cindy~L O’Bryant}, \bibinfo{person}{Davy Cheng}, \bibinfo{person}{Tristan~J Dow}, \bibinfo{person}{Bonnie Ky}, \bibinfo{person}{C~Michael Stein}, \bibinfo{person}{Anne~P Spencer}, \bibinfo{person}{Robin~J Trupp}, {and} \bibinfo{person}{JoAnn Lindenfeld}.} \bibinfo{year}{2016}\natexlab{}.
\newblock \showarticletitle{Drugs that may cause or exacerbate heart failure: a scientific statement from the American Heart Association}.
\newblock \bibinfo{journal}{\emph{Circulation}} \bibinfo{volume}{134}, \bibinfo{number}{6} (\bibinfo{year}{2016}), \bibinfo{pages}{e32--e69}.
\newblock


\bibitem[\protect\citeauthoryear{Peters, Janzing, and Sch{\"o}lkopf}{Peters et~al\mbox{.}}{2017}]%
        {peters2017elements}
\bibfield{author}{\bibinfo{person}{Jonas Peters}, \bibinfo{person}{Dominik Janzing}, {and} \bibinfo{person}{Bernhard Sch{\"o}lkopf}.} \bibinfo{year}{2017}\natexlab{}.
\newblock \bibinfo{booktitle}{\emph{Elements of Causal Inference: Foundations and Learning Algorithms}}.
\newblock \bibinfo{publisher}{MIT Press}, \bibinfo{address}{Cambridge, MA}.
\newblock
\showISBNx{978-0-262-03731-0}


\bibitem[\protect\citeauthoryear{Roy, Lin, and Ryan}{Roy et~al\mbox{.}}{2003}]%
        {roy2003scaled}
\bibfield{author}{\bibinfo{person}{Jason Roy}, \bibinfo{person}{Xihong Lin}, {and} \bibinfo{person}{Louise~M Ryan}.} \bibinfo{year}{2003}\natexlab{}.
\newblock \showarticletitle{Scaled marginal models for multiple continuous outcomes}.
\newblock \bibinfo{journal}{\emph{Biostatistics}} \bibinfo{volume}{4}, \bibinfo{number}{3} (\bibinfo{year}{2003}), \bibinfo{pages}{371--383}.
\newblock


\bibitem[\protect\citeauthoryear{Rubin}{Rubin}{2005}]%
        {rubin2005causal}
\bibfield{author}{\bibinfo{person}{Donald~B Rubin}.} \bibinfo{year}{2005}\natexlab{}.
\newblock \showarticletitle{Causal inference using potential outcomes: Design, modeling, decisions}.
\newblock \bibinfo{journal}{\emph{Journal of the American Statistical Association}} \bibinfo{volume}{100}, \bibinfo{number}{469} (\bibinfo{year}{2005}), \bibinfo{pages}{322--331}.
\newblock


\bibitem[\protect\citeauthoryear{Sandercock, Niewada, Członkowska, and Group}{Sandercock et~al\mbox{.}}{2011}]%
        {sandercock2011international}
\bibfield{author}{\bibinfo{person}{Peter~AG Sandercock}, \bibinfo{person}{Maciej Niewada}, \bibinfo{person}{Anna Członkowska}, {and} \bibinfo{person}{International Stroke Trial~Collaborative Group}.} \bibinfo{year}{2011}\natexlab{}.
\newblock \showarticletitle{The international stroke trial database}.
\newblock \bibinfo{journal}{\emph{Trials}} \bibinfo{volume}{12}, \bibinfo{number}{1} (\bibinfo{year}{2011}), \bibinfo{pages}{101}.
\newblock


\bibitem[\protect\citeauthoryear{Shalit, Johansson, and Sontag}{Shalit et~al\mbox{.}}{2017}]%
        {shalit2017estimating}
\bibfield{author}{\bibinfo{person}{Uri Shalit}, \bibinfo{person}{Fredrik~D. Johansson}, {and} \bibinfo{person}{David Sontag}.} \bibinfo{year}{2017}\natexlab{}.
\newblock \showarticletitle{Estimating individual treatment effect: Generalization bounds and algorithms}. In \bibinfo{booktitle}{\emph{International Conference on Machine Learning}}.
\newblock


\bibitem[\protect\citeauthoryear{Sohl-Dickstein, Weiss, Maheswaranathan, and Ganguli}{Sohl-Dickstein et~al\mbox{.}}{2015}]%
        {sohl2015deep}
\bibfield{author}{\bibinfo{person}{Jascha Sohl-Dickstein}, \bibinfo{person}{Eric Weiss}, \bibinfo{person}{Niru Maheswaranathan}, {and} \bibinfo{person}{Surya Ganguli}.} \bibinfo{year}{2015}\natexlab{}.
\newblock \showarticletitle{Deep unsupervised learning using nonequilibrium thermodynamics}. In \bibinfo{booktitle}{\emph{International Conference on Machine Learning}}.
\newblock


\bibitem[\protect\citeauthoryear{Song and Ermon}{Song and Ermon}{2019}]%
        {song2019generative}
\bibfield{author}{\bibinfo{person}{Yang Song} {and} \bibinfo{person}{Stefano Ermon}.} \bibinfo{year}{2019}\natexlab{}.
\newblock \showarticletitle{Generative modeling by estimating gradients of the data distribution}. In \bibinfo{booktitle}{\emph{Advances in Neural Information Processing Systems}}.
\newblock


\bibitem[\protect\citeauthoryear{Song, Sohl-Dickstein, Kingma, Kumar, Ermon, and Poole}{Song et~al\mbox{.}}{2020}]%
        {song2020score}
\bibfield{author}{\bibinfo{person}{Yang Song}, \bibinfo{person}{Jascha Sohl-Dickstein}, \bibinfo{person}{Diederik~P Kingma}, \bibinfo{person}{Abhishek Kumar}, \bibinfo{person}{Stefano Ermon}, {and} \bibinfo{person}{Ben Poole}.} \bibinfo{year}{2020}\natexlab{}.
\newblock \showarticletitle{Score-based generative modeling through stochastic differential equations}. In \bibinfo{booktitle}{\emph{International Conference on Learning Representations}}.
\newblock


\bibitem[\protect\citeauthoryear{Spiegelhalter}{Spiegelhalter}{2017}]%
        {spiegelhalter2017risk}
\bibfield{author}{\bibinfo{person}{David Spiegelhalter}.} \bibinfo{year}{2017}\natexlab{}.
\newblock \showarticletitle{Risk and uncertainty communication}.
\newblock \bibinfo{journal}{\emph{Annual Review of Statistics and Its Application}} \bibinfo{volume}{4}, \bibinfo{number}{1} (\bibinfo{year}{2017}), \bibinfo{pages}{31--60}.
\newblock


\bibitem[\protect\citeauthoryear{Teixeira-Pinto and Mauri}{Teixeira-Pinto and Mauri}{2011}]%
        {teixeira2011statistical}
\bibfield{author}{\bibinfo{person}{Armando Teixeira-Pinto} {and} \bibinfo{person}{Laura Mauri}.} \bibinfo{year}{2011}\natexlab{}.
\newblock \showarticletitle{Statistical analysis of noncommensurate multiple outcomes}.
\newblock \bibinfo{journal}{\emph{Circulation: Cardiovascular Quality and Outcomes}} \bibinfo{volume}{4}, \bibinfo{number}{6} (\bibinfo{year}{2011}), \bibinfo{pages}{650--656}.
\newblock


\bibitem[\protect\citeauthoryear{Thai, Solomon, Sequist, Gainor, and Heist}{Thai et~al\mbox{.}}{2021}]%
        {thai2021lung}
\bibfield{author}{\bibinfo{person}{Alesha~A. Thai}, \bibinfo{person}{Benjamin~J. Solomon}, \bibinfo{person}{Lecia~V. Sequist}, \bibinfo{person}{Justin~F. Gainor}, {and} \bibinfo{person}{Rebecca~S. Heist}.} \bibinfo{year}{2021}\natexlab{}.
\newblock \showarticletitle{{Lung cancer}}.
\newblock \bibinfo{journal}{\emph{The Lancet}} \bibinfo{volume}{398}, \bibinfo{number}{10299} (\bibinfo{year}{2021}), \bibinfo{pages}{535--554}.
\newblock


\bibitem[\protect\citeauthoryear{Van Der~Bles, Van Der~Linden, Freeman, Mitchell, Galvao, Zaval, and Spiegelhalter}{Van Der~Bles et~al\mbox{.}}{2019}]%
        {van2019communicating}
\bibfield{author}{\bibinfo{person}{Anne~Marthe Van Der~Bles}, \bibinfo{person}{Sander Van Der~Linden}, \bibinfo{person}{Alexandra~LJ Freeman}, \bibinfo{person}{James Mitchell}, \bibinfo{person}{Ana~B Galvao}, \bibinfo{person}{Lisa Zaval}, {and} \bibinfo{person}{David~J Spiegelhalter}.} \bibinfo{year}{2019}\natexlab{}.
\newblock \showarticletitle{Communicating uncertainty about facts, numbers and science}.
\newblock \bibinfo{journal}{\emph{Royal Society Open Science}} \bibinfo{volume}{6}, \bibinfo{number}{5} (\bibinfo{year}{2019}), \bibinfo{pages}{181870}.
\newblock


\bibitem[\protect\citeauthoryear{Wager and Athey}{Wager and Athey}{2018}]%
        {wager2018estimation}
\bibfield{author}{\bibinfo{person}{Stefan Wager} {and} \bibinfo{person}{Susan Athey}.} \bibinfo{year}{2018}\natexlab{}.
\newblock \showarticletitle{Estimation and inference of heterogeneous treatment effects using random forests}.
\newblock \bibinfo{journal}{\emph{Journal of the American Statistical Association}} \bibinfo{volume}{113}, \bibinfo{number}{523} (\bibinfo{year}{2018}), \bibinfo{pages}{1228--1242}.
\newblock


\bibitem[\protect\citeauthoryear{Wang, McDermott, Chauhan, Ghassemi, Hughes, and Naumann}{Wang et~al\mbox{.}}{2020}]%
        {wang2020mimic}
\bibfield{author}{\bibinfo{person}{Shirly Wang}, \bibinfo{person}{Matthew~BA McDermott}, \bibinfo{person}{Geeticka Chauhan}, \bibinfo{person}{Marzyeh Ghassemi}, \bibinfo{person}{Michael~C Hughes}, {and} \bibinfo{person}{Tristan Naumann}.} \bibinfo{year}{2020}\natexlab{}.
\newblock \showarticletitle{Mimic-extract: A data extraction, preprocessing, and representation pipeline for MIMIC-III}. In \bibinfo{booktitle}{\emph{ACM Conference on Health, Inference, and Learning}}.
\newblock


\bibitem[\protect\citeauthoryear{Weberpals, Feuerriegel, van~der Schaar, and Kehl}{Weberpals et~al\mbox{.}}{2025}]%
        {Weberpals2025}
\bibfield{author}{\bibinfo{person}{Janick Weberpals}, \bibinfo{person}{Stefan Feuerriegel}, \bibinfo{person}{Mihaela van~der Schaar}, {and} \bibinfo{person}{Kenneth~L. Kehl}.} \bibinfo{year}{2025}\natexlab{}.
\newblock \showarticletitle{Opportunities for causal machine learning in precision oncology}.
\newblock \bibinfo{journal}{\emph{NEJM AI}} (\bibinfo{year}{2025}).
\newblock


\bibitem[\protect\citeauthoryear{Wu, Liu, Yan, Fu, Wang, Wang, and Sun}{Wu et~al\mbox{.}}{2023}]%
        {wu2023blessings}
\bibfield{author}{\bibinfo{person}{Yong Wu}, \bibinfo{person}{Mingzhou Liu}, \bibinfo{person}{Jing Yan}, \bibinfo{person}{Yanwei Fu}, \bibinfo{person}{Shouyan Wang}, \bibinfo{person}{Yizhou Wang}, {and} \bibinfo{person}{Xinwei Sun}.} \bibinfo{year}{2023}\natexlab{}.
\newblock \showarticletitle{The blessings of multiple treatments and outcomes in treatment effect estimation}.
\newblock \bibinfo{journal}{\emph{arXiv preprint}} (\bibinfo{year}{2023}).
\newblock


\bibitem[\protect\citeauthoryear{Yao, Lo, Nir, Tan, Evnine, Lerer, and Peysakhovich}{Yao et~al\mbox{.}}{2022}]%
        {yao2022efficient}
\bibfield{author}{\bibinfo{person}{Leon Yao}, \bibinfo{person}{Caroline Lo}, \bibinfo{person}{Israel Nir}, \bibinfo{person}{Sarah Tan}, \bibinfo{person}{Ariel Evnine}, \bibinfo{person}{Adam Lerer}, {and} \bibinfo{person}{Alex Peysakhovich}.} \bibinfo{year}{2022}\natexlab{}.
\newblock \showarticletitle{Efficient heterogeneous treatment effect estimation with multiple experiments and multiple outcomes}.
\newblock \bibinfo{journal}{\emph{arXiv preprint arXiv:2206.04907}} (\bibinfo{year}{2022}).
\newblock


\bibitem[\protect\citeauthoryear{Yoon, Fitzmaurice, Lipsitz, Horton, Laird, and Normand}{Yoon et~al\mbox{.}}{2011}]%
        {yoon2011alternative}
\bibfield{author}{\bibinfo{person}{Frank~B Yoon}, \bibinfo{person}{Garrett~M Fitzmaurice}, \bibinfo{person}{Stuart~R Lipsitz}, \bibinfo{person}{Nicholas~J Horton}, \bibinfo{person}{Nan~M Laird}, {and} \bibinfo{person}{Sharon-Lise~T Normand}.} \bibinfo{year}{2011}\natexlab{}.
\newblock \showarticletitle{Alternative methods for testing treatment effects on the basis of multiple outcomes: simulation and case study}.
\newblock \bibinfo{journal}{\emph{Statistics in medicine}} \bibinfo{volume}{30}, \bibinfo{number}{16} (\bibinfo{year}{2011}), \bibinfo{pages}{1917--1932}.
\newblock


\bibitem[\protect\citeauthoryear{Yoon, Jordon, and {van der Schaar}}{Yoon et~al\mbox{.}}{2018}]%
        {yoon2018ganite}
\bibfield{author}{\bibinfo{person}{Jinsung Yoon}, \bibinfo{person}{James Jordon}, {and} \bibinfo{person}{Mihaela {van der Schaar}}.} \bibinfo{year}{2018}\natexlab{}.
\newblock \showarticletitle{{GANITE}: Estimation of individualized treatment effects using generative adversarial nets}. In \bibinfo{booktitle}{\emph{International Conference on Learning Representations}}.
\newblock


\bibitem[\protect\citeauthoryear{Zampieri, Casey, Shankar-Hari, Harrell~Jr, and Harhay}{Zampieri et~al\mbox{.}}{2021}]%
        {zampieri2021using}
\bibfield{author}{\bibinfo{person}{Fernando~G Zampieri}, \bibinfo{person}{Jonathan~D Casey}, \bibinfo{person}{Manu Shankar-Hari}, \bibinfo{person}{Frank~E Harrell~Jr}, {and} \bibinfo{person}{Michael~O Harhay}.} \bibinfo{year}{2021}\natexlab{}.
\newblock \showarticletitle{Using Bayesian methods to augment the interpretation of critical care trials. An overview of theory and example reanalysis of the alveolar recruitment for acute respiratory distress syndrome trial}.
\newblock \bibinfo{journal}{\emph{American Journal of Respiratory and Critical Care Medicine}} \bibinfo{volume}{203}, \bibinfo{number}{5} (\bibinfo{year}{2021}), \bibinfo{pages}{543--552}.
\newblock


\bibitem[\protect\citeauthoryear{Zhang, Fang, Wu, and Yu}{Zhang et~al\mbox{.}}{2024a}]%
        {zhang2024diffusion}
\bibfield{author}{\bibinfo{person}{Hengrui Zhang}, \bibinfo{person}{Liancheng Fang}, \bibinfo{person}{Qitian Wu}, {and} \bibinfo{person}{Philip~S Yu}.} \bibinfo{year}{2024}\natexlab{a}.
\newblock \showarticletitle{Diffusion-nested auto-regressive synthesis of heterogeneous tabular data}.
\newblock \bibinfo{journal}{\emph{arXiv preprint arXiv:2410.21523}} (\bibinfo{year}{2024}).
\newblock


\bibitem[\protect\citeauthoryear{Zhang, Zhang, Srinivasan, Shen, Qin, Faloutsos, Rangwala, and Karypis}{Zhang et~al\mbox{.}}{2024b}]%
        {zhang2023mixed}
\bibfield{author}{\bibinfo{person}{Hengrui Zhang}, \bibinfo{person}{Jiani Zhang}, \bibinfo{person}{Balasubramaniam Srinivasan}, \bibinfo{person}{Zhengyuan Shen}, \bibinfo{person}{Xiao Qin}, \bibinfo{person}{Christos Faloutsos}, \bibinfo{person}{Huzefa Rangwala}, {and} \bibinfo{person}{George Karypis}.} \bibinfo{year}{2024}\natexlab{b}.
\newblock \showarticletitle{Mixed-type tabular data synthesis with score-based diffusion in latent space}. In \bibinfo{booktitle}{\emph{International Conference on Learning Representations}}.
\newblock


\end{thebibliography}

\appendix
\onecolumn

\section{Dataset}
\label{app:all_dataset}

\subsection{Synthetic dataset}
\label{app:syn-data}

We construct our synthetic dataset using covariates $X$ from the ACIC 2018 \cite{macdorman1998infant} dataset with $d_X = 177$. Covariates of ACIC 2018 are derived from the linked birth and infant death data. The data generation process is described as follows,

\begin{equation}
    \begin{cases}
        A\sim \mathrm{Bernoulli}(0.5), \\ 
        Y^1=f_1(X, A)+\epsilon_1 , \\
        Y^2=\rho \cdot f_1(X, A)+(1-\rho) \cdot f_2(X, A)+\epsilon_2, \\
        f_1(X, A)=\operatorname{ReLU}(\beta_1^T X+ (\delta_1^T X) A ) + \gamma_1 A+\gamma_2 A^2+\gamma_3 \sin (A) + \alpha_1 \cdot\|X\|^2+\alpha_2 \cdot \exp \left(X^T \theta_1\right)+\alpha_3 \cdot \cos \left(X^T \theta_2 \right), \\
        f_2(X, A)=\beta_2^T X+\xi_1 A+\xi_2 A^2+\xi_3 \sin (A)+(\delta_2^T X) A + \eta_1 \cdot \log (1+|X|) + \eta_2 \cdot\|X\|^2+ \eta_3 \cdot \exp (X^T \phi_1)   
    \end{cases}
\end{equation}
where $\rho \in[0,1]$ is a hyperparameter controlling the correlation between $Y^1$ and $Y^2$, and $\epsilon_1, \epsilon_2 \sim \mathcal{N}\left(0, \sigma^2\right)$ are independent noise terms, with $\sigma = 1$. Both $f_1(X, A)$ and $f_2(X, A)$ are deterministic functions of $X$ and $A$, where $\beta, \gamma, \xi, \delta, \eta$ are randomly generated coefficients. We sample $100,000$ observations and use 20\% as the test set.

The parameter $\rho$ governs the correlation strength between $Y^1$ and $Y^2$. Specifically, when $\rho=1$, $Y^2$ is fully correlated with $Y^1$ (as $Y^2 \approx f_1(X, A)$ ). Conversely, when $\rho=0, Y^2$ becomes independent of $Y^1$ (as $Y^2 \approx f_2(X, A)$ ).

\subsubsection{Bivariate normal distribution} 
We simulate random variables $ Y^1 $ and $ Y^2 $ from the conditional bivariate normal distribution $\mathcal{N}(\mu(X, A), \Sigma(X, A))$, where the mean vector and covariance matrix are given by:
\begin{equation}
\mu(X, A) = 
\begin{bmatrix} 
\mu_1(X, A) \\ 
\mu_2(X, A) 
\end{bmatrix}, \quad 
\Sigma(X, A) = 
\begin{bmatrix} 
\sigma_1^2 & \rho \sigma_1 \sigma_2 \\ 
\rho \sigma_1 \sigma_2 & \sigma_2^2 
\end{bmatrix}.
\end{equation}

The conditional means, variances, and correlation coefficient are parameterized as:
\begin{align}
\begin{cases}
\mu_1(X, A) = \beta_{1} X^2 + \beta_{2} A + \beta_{3} \cdot (X^T A), \\
\mu_2(X, A) = \gamma_{1} X + \gamma_{2} A + \gamma_{3} \cdot (X^T A) + \gamma_{4} \cdot \log(1 + |X|) A, \\
\sigma_1^2 = \exp(\delta_{1} X + \delta_{2} A), \\
\sigma_2^2 = \exp(\delta_{3} X + \delta_{4} A), \\
\rho = \tanh(\eta_{1} X + \eta_{2} A).
\end{cases}
\end{align}
where the coefficients $\beta, \gamma, \delta, \eta$ are randomly sampled from $\mathcal{U}(-1,1)$. The conditional correlation coefficient, $\rho$, lies within $(-1,1)$, where $\rho=1$ indicates perfect positive correlation, and $\rho=0$ implies independence.

\subsection{ACIC 2018 datasets}
\label{app:acic-data}
ACIC2018 \cite{macdorman1998infant} contains 24 different settings of benchmark datasets. They are designed to benchmark causal inference algorithms with various data-generating mechanisms. Covariates of ACIC 2018 are derived from the linked birth and infant death data. ACIC 2018 provides 63 distinct data-generating mechanisms with around 40 non-equal-sized samples for each mechanism ($n$ ranges from $1,000$ to $50,000$, $d_X = 177$).

\subsection{The International Stroke Trial database}
The International Stroke Trial (IST) \cite{sandercock2011international} was conducted between 1991 and 1996. It is one of the largest randomized controlled trials in acute stroke treatment. The dataset comprises $19,435$ patients from $467$ hospitals across $36$ countries, enrolled within $48$ hours of stroke onset. The treatment assignments include trial aspirin allocation and trial heparin allocation. The trial aimed to provide reliable evidence of the safety and efficacy of aspirin and subcutaneous heparin in the clinical course of acute ischaemic stroke. The outcomes are the probability of new strokes due to blocked vessels and death or dependency at 6 months. Recurrent strokes may result in death and requiring assistance for daily activities. Covariates include age, sex, presence of atrial fibrillation, systolic blood pressure, infarct visibility on CT, prior heparin use, prior aspirin use, and recorded deficits of different body parts.

\subsection{MIMIC-III dataset}

The Medical Information Mart for Intensive Care (MIMIC-III) \cite{johnson2016mimic} is a large, single-center database comprising information relating to patients admitted to critical care units at a large tertiary care hospital. MIMIC-III contains $38,597$ distinct adult patients. We follow the standardized preprocessing pipeline \cite{wang2020mimic} of the MIMIC-III dataset. We use demographic variables such as gender and age, along with clinical variables including vital signs and laboratory test results upon admission, as confounders. These include glucose, hematocrit, creatinine, sodium, blood urea nitrogen, hemoglobin, heart rate, mean blood pressure, platelets, respiratory rate, bicarbonate, red blood cell count, and anion gap, among others. The treatment is mechanical ventilation. The outcomes are the length of hospital stays and long-term mortality outcome (90-day mortality). Survivors typically had shorter hospital stays. 

\section{Implementation details}
\label{app:implementation_details}

\subsection{Implementation details of our method}
\label{app:implentation_details_ours}

We begin by converting both continuous and discrete features into a uniform embedding space. For the continuous features, we employ a two-layer multilayer perceptron (MLP) that processes the input features through two hidden layers before mapping them to the final embedding dimension. Throughout this network, we use the SiLU activation function to introduce nonlinearity. For each discrete feature, we assign a unique, learnable embedding vector to every possible category, thereby ensuring that all features share the same embedding dimension. Once the features have been embedded, we use positional encoding, and the embedding further is processed using a series of transformer blocks. We adopt a backbone similar to the Vision Transformer (ViT). Each block integrates a multi-head self-attention mechanism with a feed-forward network. For our experiments, we stack six such transformer blocks, each employing four attention heads. We predict the distributions of the corresponding outcome conditioned on the learned representation through the transformer layers. For each discrete feature outcome, we use a two-layer MLP as the predictor. This predictor takes the output embedding for the feature and processes it through a hidden layer again using the SiLU activation function.

\textbf{Architecture of diffusion model:} Our approach employs a diffusion model in which the denoising neural network is built using a multilayer perceptron (MLP). The model architecture follows \cite{zhang2024diffusion}. Before feeding data into the diffusion process, we first convert all inputs into a shared embedding space. There are three types of inputs: (1)~Continuous data: The primary continuous features are projected into a fixed diffusion embedding space through a straightforward linear transformation. (2)~Conditional information: The additional conditional features are also mapped into the same embedding space using another linear transformation. (3)~Time step: The diffusion time step is encoded using a sinusoidal embedding scheme that produces a vector matching the diffusion embedding dimension. After these transformations, the outputs from the three encoders -- the projection of the continuous data, the projected conditional information, and the sinusoidal time embedding—are summed together. This combined representation serves as the input to the denoising network. The denoising network is implemented as a four-layer MLP. Each layer uses a SiLU activation function. The network architecture is designed to initially expand the input representation to a higher-dimensional space (specifically, twice the size of the diffusion embedding dimension) in the first two hidden layers, and then reduce it back to the original embedding dimension in the final output layer. 

\textbf{Training time:} Training our diffusion models is computationally cheap as we use a lightweight denoising network. Let $K$ be the number of outcomes, $N$ the feature dimensions of confounding ($K \ll N$), and $d$ be the hidden dimensions of NN. The main cost comes from (1)~transformer model: self-attention $O(N^2 * d)$ and feed-forward network $O(N * d^2)$. (2)~Conditional diffusion models: $O(K*d^2)$ for each predicted outcome. Since the diffusion part uses a lightweight shallow MLP, its cost is negligible compared to the transformer. Empirically, training \methodshort takes about 1 hour on an NVIDIA A100 GPU.

\subsection{Implementation details of baselines}
\label{app:baseline_implementation} 

We follow the implementation from \url{https://github.com/AliciaCurth/CATENets/tree/main} for most of the CATE estimators, including S-Net \cite{kunzel2019metalearners}, T-Net \cite{kunzel2019metalearners}, TARNet \cite{shalit2017estimating}, CFRNet \cite{shalit2017estimating}. For GANITE \cite{yoon2018ganite}, we follow the implementation of \url{https://github.com/vanderschaarlab/mlforhealthlabpub/tree/main/alg/ganite}. For DiffPO \cite{ma2024diffpo}, we follow the implementation of \url{https://github.com/yccm/DiffPO}. We performed hyperparameters tuning of the nuisance functions models for all the baselines based on five-fold cross-validation using the training subset. For each baseline, we performed a grid search with respect to different tuning criteria, evaluated on the validation subsets. For evaluating the uncertainty estimation, for both training and testing, the dropout probabilities were set to $p = 0.1$.

\end{document}